\documentclass[runningheads]{llncs}
\usepackage{graphicx}
\usepackage{tikz}
\usepackage{comment}
\usepackage{amsmath,amssymb} 
\usepackage{color}
\usepackage{multirow}
\usepackage{subfigure}
\usepackage{url}
\usepackage{booktabs} 
\usepackage{multirow}
\usepackage[rightcaption]{sidecap}
\usepackage{hyperref}

\begin{document}
\pagestyle{headings}
\mainmatter

\title{Rethinking Pseudo-LiDAR Representation}
\titlerunning{Rethinking Pseudo-LiDAR Representation} 
\authorrunning{X. Ma et al.} 
\author{Xinzhu Ma\inst{1} \and
        Shinan Liu\inst{2} \and
        Zhiyi Xia\inst{3} \and
        Hongwen Zhang\inst{4} \and \\
        Xingyu Zeng\inst{2} \and
        Wanli Ouyang\inst{1}}

\institute{The University of Sydney, SenseTime Computer Vision Research Group, Australia \\
\email{\{xima0693@uni., wanli.ouyang@\}sydney.edu.au} \and
SenseTime Research, China \\
\email{\{liushinan, zengxingyu\}@sensetime.com} \and
Dalian University of Technology, China \\
\email{xiazhiyi99@mail.dlut.edu.cn}  \and 
Institute of Automation, Chinese Academy of Sciences, China \\
\email{hongwen.zhang@cripac.ia.ac.cn}}

\maketitle


\begin{abstract}
The recently proposed pseudo-LiDAR based 3D detectors greatly improve the benchmark of monocular/stereo 3D detection task.
However, the underlying mechanism remains obscure to the research community.
In this paper, we perform an in-depth investigation and observe that the efficacy of pseudo-LiDAR representation comes from the coordinate transformation, instead of data representation itself.
Based on this observation, we design an image based CNN detector named PatchNet, which is more generalized and can be instantiated as pseudo-LiDAR based 3D detectors.
Moreover, the pseudo-LiDAR data in our PatchNet is organized as the image representation, which means existing 2D CNN designs can be easily utilized for extracting deep features from input data and boosting 3D detection performance.
We conduct extensive experiments on the challenging KITTI dataset, where the proposed PatchNet outperforms all existing pseudo-LiDAR based counterparts.
Code has been made available at: \url{https://github.com/xinzhuma/patchnet}.

\end{abstract}


\section{Introduction}

3D object detection has received increasing attention from both industry and academia because of its wide applications in various fields such as autonomous driving and robotics.
Existing algorithms largely rely on LiDAR sensors, which provide accurate 3D point clouds of the surrounding environment.
Although these approaches achieve impressive performance, the excessive dependence on expensive equipment restricts their application prospects.

Compared with fast developing LiDAR-based algorithms, 3D detection~\cite{chen20153d,chen2016monocular,Li_2019_CVPR} results produced from only RGB images lag considerably behind.
This can be attributed to the ill-posed nature of the problem, where a lack of explicit knowledge about the unobserved depth dimension significantly increases the task complexity.
An intuitive solution is that we can use a Convolutional Neural Network (CNN) to predict the depth map~\cite{Alhashim2018,fu2018deep,godard2017unsupervised} and then use it to augment the input data if we do not have the available depth information. 
Although the estimated depth map is helpful to 3D scene understanding, the performance improvement brought by it is still limited.

Several recently proposed algorithms~\cite{Wang_2019_CVPR,Ma_2019_ICCV,Weng_2019_ICCV_Workshops} transform the estimated depth map into pseudo-LiDAR representation, and then apply LiDAR based methods to the transformed data.
Surprisingly, this simple yet effective method achieves significant improvement in detection accuracy on the challenging KITTI dataset.
However, it is unclear why such a representation can bring so much performance improvement. 
According to the empirical explanation of proponents, the choice of representations is the critical success factor of 3D detection systems.
Compared with image representation, they believe that pseudo-LiDAR is more suitable for describing the 3D structure of objects, which is the main reason for  performance improvement.
However, in the absence of direct evidence, the correctness of this statement is still open to doubt.

In this paper, we aim to explore the essential reasons of this phenomenon.
Specifically, on the basis of prior works, we carefully construct an image representation based detector named PatchNet-vanilla, which is an equivalent implementation of pseudo-LiDAR~\cite{Wang_2019_CVPR} except for the representation of input data.
With this detector, we can compare the influence of these two kinds of representations on 3D detection task in depth.
Different from the arguments of other works~\cite{Wang_2019_CVPR,Ma_2019_ICCV,Weng_2019_ICCV_Workshops}, we observe that the performances of PatchNet-vanilla and pseudo-LiDAR~\cite{Wang_2019_CVPR} are completely matched, which means that data representation has no effect on 3D detection performance.
Moreover, we perform ablation studies on the input data, and observe that the real thing matters is coordinate transformation from image coordinate system to the LiDAR coordinate system, which implicitly encodes the camera calibration information into input data.

PatchNet-vanilla also hints us that pseudo-LiDAR representation is not necessary to improve the accuracy of image based 3D detection.
By integrating the generated 3D coordinates as additional channels of input data, our 3D detector gets promising performance.
More importantly, this approach can be easily generalized to other image based detectors.
Also notice that, as a kind of non-grid structured data, pseudo-LiDAR signals commonly need point-wise CNNs~\cite{qi2017pointnet,qi2017pointnet++} to process. 
However, the development of these technologies still lags behind the standard CNNs.
From this point of view, the image-based detectors should outperform their counterparts based on pseudo-LiDAR.
To confirm this hypothesis, PatchNet was proposed by extending our original model (e.g., using more powerful backbone network \cite{He_2016_CVPR,Hu_2018_CVPR}), and outperforms other pseudo-LiDAR based detectors on KITTI dataset.
In addition, there are other benefits from using images directly as the network's inputs, such as allowing us to train an end-to-end 3D detector.
Based on above reasons, we argue that image representation based 3D detectors have greater development potential.

To summarize, the contributions of this paper are as follows:
First, through sufficient experimental demonstration, we confirm that the reason why the pseudo-LiDAR representation is effective is not the data representation itself, but the coordinate system transformation.
Second, we find that pseudo-LiDAR representation is not necessary to improve detection performance.
After integrating spatial coordinates, image representation based algorithms can also achieve the competitive if not superior the same performance.
Third, thanks to more powerful image-based deep learning technologies, we achieve the state-of-the-art performance and show the potential of image representation based 3D detectors.


\section{Related Work}
\subsection{3D detectors based on image representation}
Most of the early works in this scope share the same paradigm with 2D detectors~\cite{girshick2015fast,ren2015faster,zhou2020cheaper,dai2016r,lin2017focal,Lin_2017_CVPR,he2017mask}.
However, estimating the 3D coordinates $(x, y, z)$ of the object center is much more complicated since there is ambiguity to locate the absolute physical position from only image appearances.
Mono3D~\cite{chen2016monocular} focus on 3D object proposals generation using prior knowledge (e.g., object size, ground plane).
Deep3DBox~\cite{mousavian20173d} introduces geometric constraints based on the fact that the 3D bounding box should fit tightly into 2D detection bounding box.
DeepMANTA~\cite{chabot2017deep} encodes 3D vehicle information using key points, since vehicles are rigid objects with well known geometry. 
Then the vehicle recognition in DeepMANTA can be considered as key points detection.
An expansion stage of ROI-10D~\cite{Manhardt_2019_CVPR} takes the advantage of depth information provided by an additional depth estimator~\cite{fu2018deep,chang2018pyramid} , which itself is learned in a self-supervised manner. 
In Multi-Fusion~\cite{Xu_2018_CVPR}, a multi-level fusion approach is proposed to exploit disparity estimation results from a pre-trained module for both the 2D box proposal generation and the 3D prediction part of their network.
MonoGRNet~\cite{qin2019monogrnet} consists of four subnetworks for progressive 3D localization and directly learning 3D information based on solely semantic cues.
MonoDIS~\cite{Simonelli_2019_ICCV} disentangles the loss for 2D and 3D detection and jointly trains these two tasks in an end-to-end manner.
M3D-RPN~\cite{Brazil_2019_ICCV} is the current state-of-the-art with image representation as input, using multiple 2D convolutions of non-shared weights to learn location-specific features for joint prediction of 2D and 3D boxes.
The above approaches utilize various prior knowledge, pre-train models or more powerful CNN designs, but they do not try to use pseudo-LiDAR data to improve their performance. 
Our work aims to improve the detection accuracy of image-based methods by extracting useful information from pseudo-LiDAR data, which is complementary to these approaches.

\subsection{3D detectors based on pseudo-LiDAR representation}
Recently, several approaches \cite{Ma_2019_ICCV,Wang_2019_CVPR,Weng_2019_ICCV_Workshops,you2019pseudo} greatly boost the performance of monocular 3D detection task.
What they have in common is that they first estimate the depth map from the input RGB image and transform it into pseudo-LiDAR (point cloud) by leveraging the camera calibration information.
Specifically, \cite{Wang_2019_CVPR} adopt off-the-shelf LiDAR-based 3D detectors~\cite{Qi_2018_CVPR,ku2018joint} to process the generated pseudo-LiDAR signals directly.
AM3D~\cite{Ma_2019_ICCV} proposes a multi-modal features fusion module to embed the complementary RGB cues into the generated pseudo-LiDAR representation.
Besides, \cite{Ma_2019_ICCV} also proposes a depth prior based background points segmentation module to avoid the problems caused by the inaccuracy of point cloud annotation. 
\cite{Weng_2019_ICCV_Workshops} proposes a 2D-3D bounding box consistency loss which can alleviate the local misalignment issue.
However, such methods rely heavily on the accuracy of depth map.
Overall, pseudo-LiDAR based detectors achieve impressive accuracy in 3D detection task, however, the underlying mechanism is still obscure to the research community.
In this paper, we perform an in-depth investigation on this issue. Besides, pseudo-LiDAR based detectors treat generated 3D data as point cloud and use PointNet for processing the point cloud, while our PatchNet organizes them as image and facilitates the use of 2D CNN for processing the data.


\section{Delving into pseudo-LiDAR representation}

\begin{figure}[t]
\begin{center}
\includegraphics[width=1.0\linewidth]{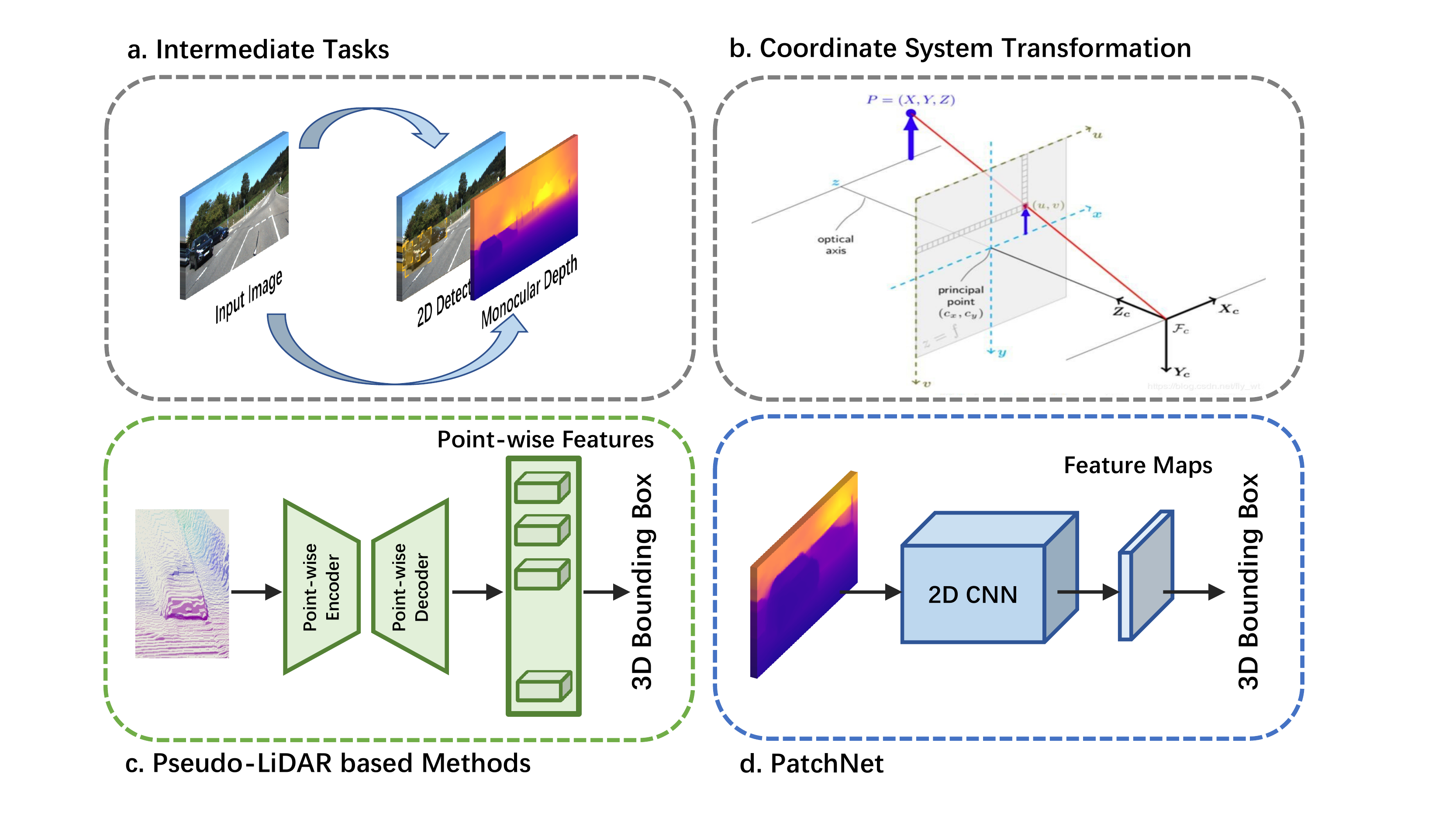}
\end{center}
\caption{{\bf Comparison of pseudo-LiDAR based methods~\cite{Ma_2019_ICCV,Wang_2019_CVPR,Weng_2019_ICCV_Workshops} and PatchNet.} They both generate intermediate tasks using off-the-shelf models (a), and project the image coordinates to the world coordinates (b).
Pseudo-LiDAR based methods treat these data as LiDAR signals, and use point-wise network to predict results from them (c).
However, PatchNet organizes them as image representation for subsequent processing (d).}
\label{fig:framework}
\end{figure}

In this section, we investigate the influence of pseudo-LiDAR representation on 3D detection accuracy.
In particular, we first give a brief review of pseudo-LiDAR based detectors and introduce the technical details of its image based equivalent detector. 
Then, we analyse whether data representation is the internal reason of performance improvement by comparing the performance of these two detectors.

\subsection{Review of pseudo-LiDAR based detectors}
Here we take pseudo-LiDAR~\cite{Wang_2019_CVPR} as example for analysis, and the paradigm of \cite{Wang_2019_CVPR} can be summarized as follows:

\noindent
{\bf Step 1: Depth estimation.}
Given a single monocular image (or stereo pairs) as input, \cite{Wang_2019_CVPR} predict the depth $d$ for each image pixel $(u, v)$ using a stand alone CNN (Fig~\ref{fig:framework}(a)).

\noindent
{\bf Step 2: 2D detection.}
Another CNN is adopted to generate 2D object region proposals (Fig~\ref{fig:framework}(a)).

\noindent
{\bf Step 3: 3D data generation.}
First, regions of interests (RoIs) are cropped from the depth map generated from Step 1, according to the region proposals generated from Step 2.
Then, the 3D coordinates of pixels of each RoI can be recovered by:
\begin{equation}
\left\{
      \begin{array}{lr}
      z = d ,\\
      x = (u-C_{x}) \times z / f  ,\\
      y = (v-C_{y}) \times z / f  ,\\
      \end{array}
\right.
\label{eq:transformation}
\end{equation}
where $f$ is the focal length of the camera, $(C_{x}, C_{y})$ is the principal point (Fig~\ref{fig:framework}(b)).

\noindent
{\bf Step 4: 3D object detection.}
Pseudo-LiDAR based approaches treat the 3D data generated from Step 3 as LiDAR signals, and use point-wise CNN to predict result from them (Fig~\ref{fig:framework}(c)).
In particular, they are treated as an unordered point set $\{x_{1},x_{2},...,x_{n}\}$ with $x_{i} \in \mathbb{R}^{d}$, and processed by PointNet, which defines a set function $f$ that maps a set of points to a output vector:
\begin{equation}
	f(x_{1},x_{2},...,x_{n}) = 
	\gamma \left( \mathop{{\bf MAX}}\limits_{i=1,...,n} \{h(x_{i})\} \right)
	\label{eq:pointnet}
\end{equation}
where $\gamma$ and $h$ are implemented by multi-layer perceptron (MLP) layers.

\subsection{PatchNet-vanilla: equivalent implementation of pseudo-LiDAR}

\subsubsection{Analysis.}
The most significant difference between the pseudo-LiDAR based approaches \cite{Ma_2019_ICCV,Wang_2019_CVPR} and other approaches lies in the representation of depth map.
The authors of \cite{Ma_2019_ICCV,Wang_2019_CVPR} argue that pseudo-LiDAR representation is more suitable for describing the 3D structure of objects, which is the main reason behind the high accuracy of their models. 
To verify this, we conduct an image representation based detector, i.e., PatchNet-vanilla, which is identical to pseudo-LiDAR~\cite{Wang_2019_CVPR} except for the input representation. 

\begin{SCfigure}
\centering
\includegraphics[width=0.48\textwidth]{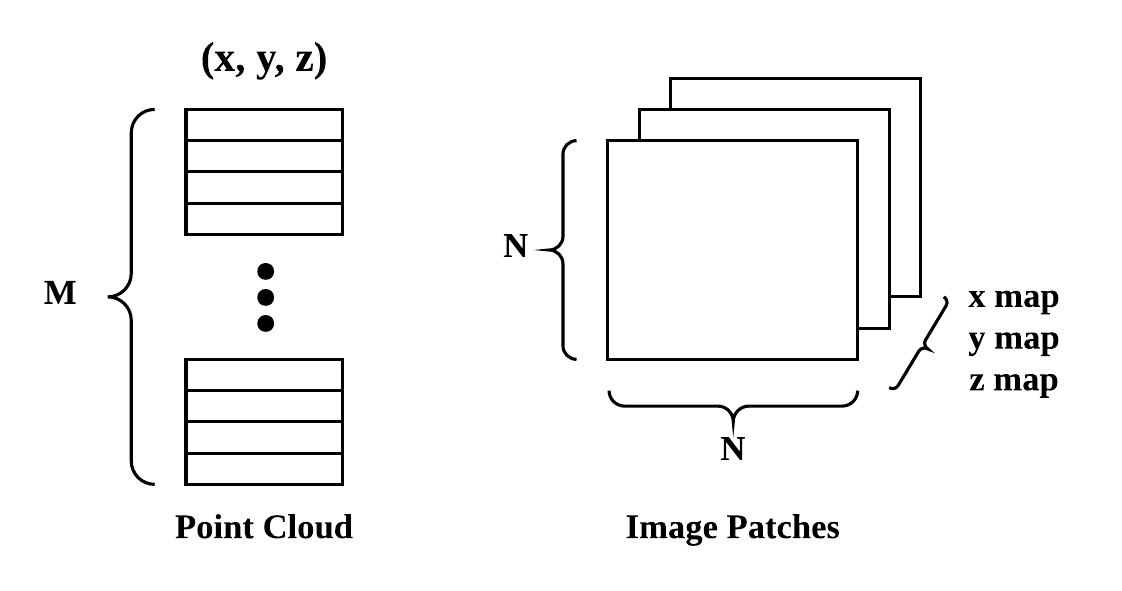}
\caption{{\bf Illustration of input data.} Pseudo-LiDAR based approaches use point cloud ({\it left}) as input, while PatchNet use image patches ({\it right}) as input. We set $M = N \times N$ so that these two kinds of input data contain the same amount of information.}
\label{fig:input}
\end{SCfigure}

\subsubsection{Implementation.}
The steps 1, 2 and 3 in PatchNet-vanilla are the same as that in the pseudo-LiDAR based detectors. 
Therefore, they have the same estimated depth, 2D detection results and generated 3D data.
The main difference is the Step 4, which will be analyzed in details.
Specifically, in PatchNet-vanilla, the generated 3D data is organized as image representation (see Fig~\ref{fig:input}), where each pixel location with 3 channels, i.e. $(x, y, z)$ in Eq.~\ref{eq:transformation}. 
Different from point-wise CNN used in pseudo-LiDAR counterparts, 2D CNN is used for processing the input data in PatchNet-vanilla (Fig~\ref{fig:framework}(d)).
Note that we can define a same function as Eq.~\ref{eq:pointnet} using 2D convolution with $1\times$1 receptive field and global max pooling.
This scheme is also adopted in the official implementation$\footnote[1]{https://github.com/charlesq34/pointnet.}$ of PointNet.

\subsection{Preliminary conclusion}
\begin{table}[!htbp]
\caption{{\bf Comparison of different input representation.}
        Experiments are conducted on KITTI {\it validation} set.
		$\ast$ indicates the method is reproduced by ourself.
		Metric is $AP|_{R_{11}}$ of the {\bf Car} category.}
\begin{center}
\begin{tabular}{lccccccc}
\toprule
\multirow{2}{*}{Method} & \multirow{2}{*}{Modality} & \multicolumn{3}{c}{3D detection} & \multicolumn{3}{c}{BEV detection} \\ 
\cmidrule{3-8} 
 ~ & ~ & Easy & Moderate & Hard  & Easy & Moderate & Hard\\ 
\midrule
pseudo-LiDAR~\cite{Wang_2019_CVPR} & pseudo-LiDAR & 28.2 & 18.5 & 16.4 & 40.6 & 26.3 & 22.9 \\ 
pseudo-LiDAR* & pseudo-LiDAR & 28.9 & 18.4 & 16.2 & 41.0 & 26.2 & 22.8  \\ 
PatchNet-vanilla & image & 28.7 & 18.4 & 16.4 & 40.8 & 26.1 & 22.8  \\  
\bottomrule
\end{tabular}
\end{center}
\label{table:comp}
\end{table}

The performances of PatchNet-vanilla and pseudo-LiDAR are reported in Tab.~\ref{table:comp}, where we reproduce pseudo-LiDAR to eliminate the impact of implementation details.
As can be seen, PatchNet-vanilla achieves almost the same accuracy as pseudo-LiDAR, which means the choice of data representation has no substantial impact on 3D detection tasks.
Moreover, we perform ablation studies on data content, and observe that coordinate transform is the key factor for performance improvement (experimental results and analysis can be found in Sec.~\ref{sec:5.2}).

Above observations reveal that pseudo-LiDAR representation is not necessary, and after integrating the generated 3D information, image representation has the same potential.
More importantly, compared with point-wise CNNs~\cite{qi2017pointnet,qi2017pointnet++}, image based representation can utilize the well-studied 2D CNNs for developing high-performance 3D detectors.
Along this direction, we show how the proposed PatchNet framework is used to further improve the detection performance in Sec.~\ref{sec:4}.


\section{PatchNet}
\label{sec:4}

\begin{figure}[t]
\begin{center}
\includegraphics[width=1.0\linewidth]{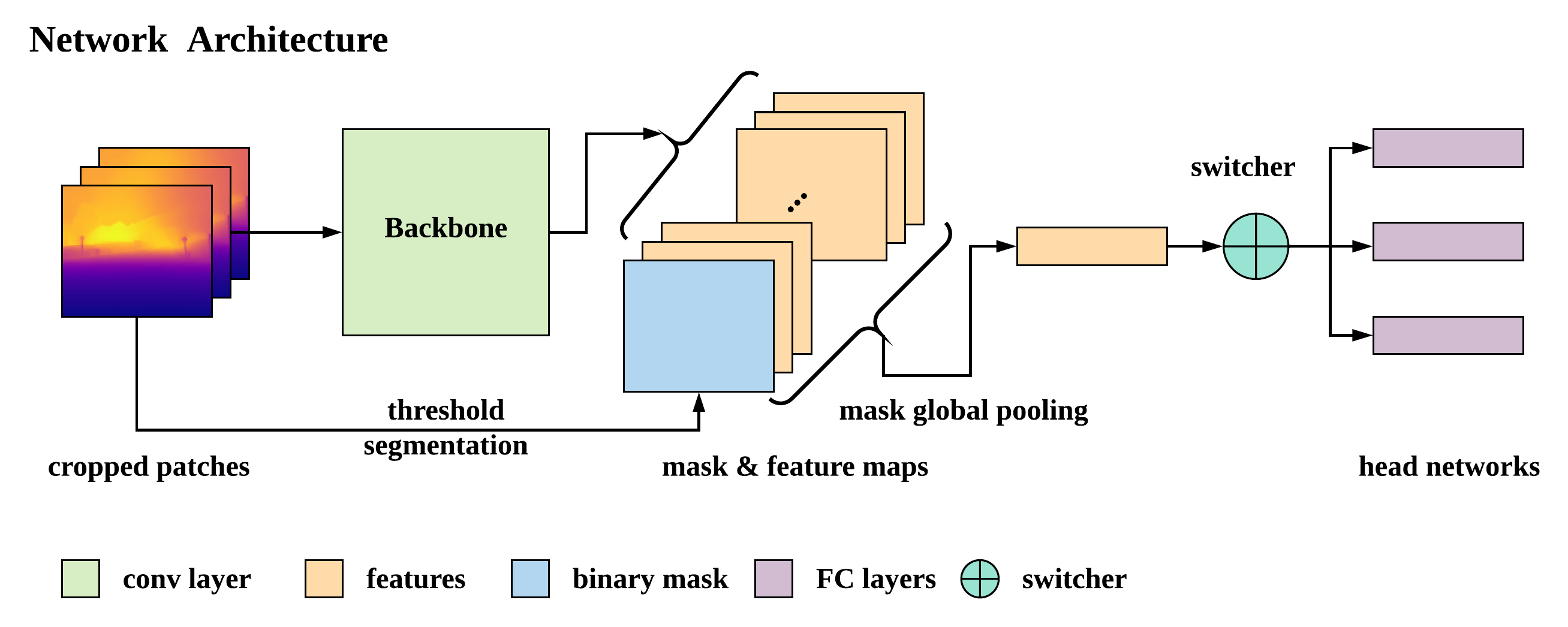}
\end{center}
   \caption{{\bf Illustration of the network architecture.} Given an input  patch with $\{x, y, z\}$ channels, we first generate a binary mask according to mean depth, and use it to guide pooling layer to extract the features corresponding to foreground object. Then, we assign examples to different head networks according to the prediction difficulty of them. }
   \label{fig:arch}
\end{figure}

In PatchNet, we first train two deep CNNs on two intermediate prediction tasks (i.e., 2D detection and depth estimation) to obtain position and depth information, which are the same as PatchNet-vanilla and pseudo-LiDAR based detectors (Fig~\ref{fig:framework}(a)).
Then, as shown in Fig~\ref{fig:arch}, for each detected 2D object proposal, we crop the corresponding region from the depth map, and recover its spatial information using Eq~\ref{eq:transformation}.
Next, deep features of RoIs are extracted by backbone network, and filtered by the mask global pooling and foreground mask.
Finally, we use a detection head with difficulty assignment mechanism to predict the 3D bounding box parameterized by $(x, y, z, h, w, l, \theta)$.

\subsubsection{Backbone}
Most of existing backbone networks can be used in our method to extract image features.
In our implementation, we use the ResNet-18~\cite{He_2016_CVPR} with Squeeze-and-Excitation (SE) block~\cite{Hu_2018_CVPR} as the 3D detection backbone. 
Moreover, we remove all pooling layers in the original SE-ResNet-18 so that its output features have the same size as input image patches.
Then we use  mask global pooling operation and generated mask to extract features from foreground object.

\subsubsection{Mask global pooling}
The feature maps $\mathbf{X}$ output from the backbone network will be converted to a feature vector by global pooling.
Conventional global pooling takes features of all positions into account and output the global feature.
To obtain more robust features, we perform global pooling only on those features within foreground regions so that the final feature is corresponding to those pixels of interest.
Specifically, we additionally generate a binary mask $\mathbf{M}$ which indicates the foreground region.
This masks will be applied to the feature maps $\mathbf{X}$ to select foreground features before global pooling.
Such a mask global pooling encourages the final feature to focus on the regions of interest.

\subsubsection{Mask generation}
Following the prior work~\cite{Ma_2019_ICCV}, 
the fore/background binary mask $\mathbf{M}$ is obtained by setting a threshold to the depth map.
Specifically, we empirically add an offset on the mean depth of each patch and set it as the threshold.
The regions with the depth values smaller than this threshold will be regarded as the foreground regions.
The binary mask $\mathbf{M}$ has the same resolution as the input image with its values corresponding to foreground regions set as 1 and otherwise 0.

\subsubsection{Head}
Inspired by difficulty-wise evaluation adopted by KITTI dataset, we use three branches to deal with samples of different difficulty levels separately. To select the branch, we need a specific module.
Specifically, before sending the feature maps to the three parallel box estimators, we add another branch to predict the difficulty level of each instance. 

Note that all three branches are the same in network architecture, and only different in learned parameters for handling different difficulty levels.
Besides, in our implementation, all three branches predict results simultaneously, and two of them are blocked according to the output of difficulty predictor.
Theoretically, this does not affect the accuracy of the algorithm, and allows all branches to run in parallel with the cost of extra GPU memory.

\subsubsection{Loss function}
The ground truth box is parameterized by center $(x, y, z)$, size $(w, h, l)$ and heading angle $\theta$.
We adopted the loss function proposed by \cite{Qi_2018_CVPR} to our baseline model:
\begin{equation}
 \mathcal{L} = \mathcal{L}_{center}+ \mathcal{L}_{size} + \mathcal{L}_{heading} + \lambda \cdot \mathcal{L}_{corner}
\end{equation}
where $\mathcal{L}_{center}, \mathcal{L}_{size}$, and $\mathcal{L}_{heading}$ respectively denote the loss function for the center, size, and heading angle.   $\lambda$ is an empirical weight, and $\mathcal{L}_{corner}$ is used to alleviate the potential sub-optimal problem.
Please refer to \cite{Qi_2018_CVPR} for details.


\section{Experiments}
\subsection{Setup}
\subsubsection{Dataset}

We evaluate our approach on the challenging KITTI dataset \cite{geiger2012we}, which provides 7,481 images for training and 7,518 images for testing. 
Detection and localization (i.e., bird's-eye-view detection) tasks are evaluated in three different subsets: {\it easy}, {\it moderate} and {\it hard}, according to the occlusion and truncation levels of objects. 
Since the ground truth for the test set is not available and the access to the test server is limited, we follow the protocol of prior works~\cite{chen2016monocular,chen20153d,Chen_2017_CVPR} to divide the training data into a training set (3,712 images) and a validation set (3,769 images).
We will conduct ablation studies based on this split and also report final results on the testing set provided by KITTI server.
Due to space limitations, we only report the {\bf Car} detection results of {\bf monocular images} in the main paper.
More results about {\bf stereo pairs} and {\bf Pedestrian/Cyclist} can be found in Appendix.

\subsubsection{Metric}
Most of previous works use 11-point interpolated average precision (IAP) metric~\cite{geiger2012we} as follows:
\begin{equation}
AP|_{R_{11}} = \frac{1}{11}\sum_{r\in R_{11}} \mathop{max}\limits_{\widetilde{r} \geq r}  \rho(\widetilde{r}).
\end{equation}
Recently, to avoid ostensible boost in performance, KITTI and \cite{Simonelli_2019_ICCV} call for a new 40-point IAP ($AP|_{R_{40}}$) with the exclusion of ``0'' and four-times denser interpolated prediction for better approximation of the area under the Precision/Recall curve.
For fair and comprehensive comparisons with previous and future works, we show both $AP|_{R_{11}}$ and $AP|_{R_{40}}$ in the following experiments.

\subsection{Investigation of pseudo-LiDAR representation}
\label{sec:5.2}
\begin{table}[!ht]
\caption{3D object detection results on KITTI {\it validation} set. 
Metrics are $AP_{3D}$ and $AP_{BEV}$  of the {\bf Car} category with 11 recall positions. $\ast$ indicates method is reproduced by ourselves.}
\begin{center}
\begin{tabular}{lccccccc}
\toprule
\multirow{2}{*}{Method} & \multirow{2}{*}{Modality} & \multicolumn{3}{c}{3D detection} & \multicolumn{3}{c}{BEV detection} \\ 
\cmidrule{3-5} \cmidrule{6-8} 
 ~ & ~ & Easy & Moderate & Hard  & Easy & Moderate & Hard\\ 
\midrule
pseudo-LiDAR~\cite{Wang_2019_CVPR} & pseudo-LiDAR & 28.2 & 18.5 & 16.4 & 40.6 & 26.3 & 22.9 \\ 
pseudo-LiDAR* & pseudo-LiDAR & 28.9 & 18.4 & 16.2 & 41.0 & 26.2 & 22.8  \\ 
AM3D~\cite{Ma_2019_ICCV} & pseudo-LiDAR & 32.2 & 21.1 & 17.3  & 43.8 & 28.4 & 23.9 \\ 
PatchNet-vanilla & image & 28.7 & 18.4 & 16.4 & 40.8 & 26.1 & 22.8  \\  
PatchNet-AM3D & image & 32.8 & 20.9 & 17.3 & 43.5 & 28.2 & 23.6 \\
PatchNet & image  & 35.1 &  22.0 &  19.6 &  44.4 &  29.1 &  24.1 \\
Improvement & -  & +2.9 & +0.9 & +2.3 & +0.6 & +0.7 & +0.2 \\
\bottomrule
\end{tabular}
\end{center}
\label{table:main_results}
\end{table}

\subsubsection{Analysis of data representation}
As shown in Tab. \ref{table:main_results}, PatchNet-vanilla shows a comparable results with pseudo-LiDAR, which indicates that {\it data representation is not the key factor to improve the performance of 3D detectors.}
To further validate this claim, we also adjust our image representation based detector based on AM3D, where we achieve a matched performance again.

\begin{table}[!ht]
\begin{center}
\caption{{\bf Comparison between different input data} on KITTI {\it validation} set. Metrics are $AP_{3D}$ and $AP_{BEV}$ of the {\bf Car} category with 11 recall positions.}
\begin{tabular}{ccccccc}
\toprule
\multirow{2}{*}{input} & \multicolumn{3}{c}{$AP_{3D}$} & \multicolumn{3}{c}{$AP_{BEV}$} \\ 
\cmidrule{2-4} \cmidrule{5-7} 
~  & Easy & Moderate & Hard  & Easy & Moderate & Hard\\ 
\midrule
 $\lbrace z \rbrace$  & 4.51 & 3.48 & 3.03& 6.31 & 4.50 & 3.98 \\
 $\lbrace x, z \rbrace$  & 27.1 & 18.3 & 15.8 & 35.9 & 23.4 & 18.3 \\ 
 $\lbrace x, y, z \rbrace$  & 35.1 &  22.0 & 19.6 & 44.4 & 29.1 & 24.1\\
 $\lbrace u, v, z \rbrace$  & 24.6 & 15.7 & 14.6 & 33.2 & 21.3 & 16.7 \\
\bottomrule
\end{tabular}
\end{center}
\label{table:inputvalues}
\end{table}

\subsubsection{Analysis of data content}
We conduct an ablation study on the effect of input channels and report the results in Tab.~\ref{table:inputvalues}. 
We can see from the results that, using only depth as an input, it is almost impossible to obtain accurate 3D bounding boxes.
If other coordinates are used, the accuracy of predicted boxes improves greatly, which validates the importance of generated spatial features.
It should be noted that in the absence of y-axis data, this detection accuracy is much worse than our full model. This is shows that all coordinates are useful for the 3D detection.

In pseudo-LiDAR, the coordinate $(u, v)$ for images is projected to the world coordinate $(x, y)$ using the camera information. Experimental results in Tab.~\ref{table:inputvalues} also compares the effectiveness of different coordinate systems.
According to experimental results, world coordinate $(x, y)$, which utilizes the camera information, performs much better than image coordinate $(u, v)$.
Through the above experiments, we can observe that \emph{that real thing matters is coordinate system transformation, instead of data representation itself}.

\subsection{Boosting the performance of PatchNet} 

\subsubsection{Backbone}
Compared with point-wise backbone nets commonly used in (pseudo) LiDAR based methods, standard 2D backbones such as \cite{He_2016_CVPR,Hu_2018_CVPR,xie2017aggregated} can extract more discriminative features, which is a natural advantage of image based detectors.
We investigate the impact of different backbones on proposed PatchNet, and the experimental results are summarized in Tab.~\ref{table:backbone} ({\it left}). The original PointNet has only 8 layers. For fair comparison, we construct a PointNet with 18 layers, which is denoted by PointNet-18 in Tab.~\ref{table:backbone}.
Compared with PointNet-18, using 2D convolution backbones can  improve the accuracy of 3D boxes, especially for {\it hard} setting.
This is because these cases are usually occluded/truncated or far away from the camera, and estimating the pose of them is more dependent on context information.
However, it is evident that the point-wise CNNs are hard to extract local features of data efficiently. 
From this perspective, image representation based detectors have greater development potentials. 
Besides, we can see from Tab.~\ref{table:backbone} ({\it right}) that the accuracy does not improve much when the CNN has more layers from ResNeXt-18 to ResNeXt-50. Compared with ResNeXt-50, ResNeXt-101 performs worse, which can be attributed to over-fitting. All the CNNs are trained from scratch. 

\begin{table}[!ht]
\caption{{\bf Comparisons of different backbone nets} on KITTI {\it validation} set. Metrics are $AP_{3D}|_{R_{11}}$ for 3D detection task of the {\bf Car} category with IoU threshold = 0.7. Other settings are same as PatchNet-vanilla.}
\begin{center}
\begin{minipage}[t]{0.48\textwidth}
\centering
\begin{tabular}{lccc}
\toprule
Backbone  & Easy & Moderate & Hard \\ 
\midrule
PointNet-18  & 31.1 & 20.5 & 17.0 \\
ResNet-18  & 33.2 & 21.3 & 19.1 \\
ResNeXt-18  & 33.4 & 21.2 & 19.2 \\
SE-ResNet-18  & 33.7 & 21.5 & 19.2 \\
\bottomrule
\end{tabular}
\end{minipage}
\begin{minipage}[t]{0.48\textwidth}
\centering
\begin{tabular}{lccc}
\toprule
Backbone  & Easy & Moderate & Hard \\ 
\midrule
ResNeXt-18 & 32.7 & 21.2 & 19.2 \\
ResNeXt-50 & 32.9 & 21.4 & 17.3 \\
ResNeXt-101 & 31.1 & 20.9 & 17.0 \\
\bottomrule
\end{tabular}
\end{minipage}
\end{center}
\label{table:backbone}
\end{table}

\subsubsection{Mask global pooling.}
In the PatchNet, we design the mask global pooling operation to force the feature maps must be extracted from a set of pixels of interests, which can be regarded as a hard attention mechanism.
Tab.~\ref{table:maskpooling} shows the effectiveness of this operation, e.g., mask global pooling (max) can improve $AP_{3D}|_{11}$ by 1.4\%  for moderate setting and by 2.7\% for easy setting, and max pooling is slightly better than avg pooling.
Besides, the visualization result shown in Fig.~\ref{table:maskpooling} intuitively explains the reason for the performance improvement.
Specifically, most activation units filtered by mask global pooling correspond to foreground goals, while the ones from standard global max pooling will have many activation units on the background.

It should be noted that the background points provide contextual information in our model, but they are not involved in \cite{Qi_2018_CVPR,Wang_2019_CVPR} as input for PointNet.

\begin{table}[!ht]
\begin{center}
\caption{{\bf Ablation study of mask global pooling} on KITTI {\it validation} set. Metrics are $AP_{3D}$ and $AP_{BEV}$ of the {\bf Car} category with 11 recall positions. Other settings are same as PatchNet (full model).}
\begin{tabular}{cccccccc}
\toprule
\multirow{2}{*}{pooling}& \multirow{2}{*}{type} & \multicolumn{3}{c}{$AP_{3D}$} & \multicolumn{3}{c}{$AP_{BEV}$} \\ 
\cmidrule{3-5} \cmidrule{6-8} 
~ & ~  & Easy & Moderate & Hard  & Easy & Moderate & Hard\\ 
\midrule
standard & max &32.4 & 20.6 & 17.7 & 41.3 & 27.0 & 21.6 \\
mask & avg & 34.6 & 21.6 & 19.3 & 43.5 & 28.7 & 23.3 \\
mask & max & 35.1 & 22.0 & 19.6 & 44.4 & 29.1 & 24.1 \\
\bottomrule
\end{tabular}
\end{center}
\label{table:assignment}
\end{table}

\begin{figure}[t]
\begin{center}
\includegraphics[width=0.2\linewidth]{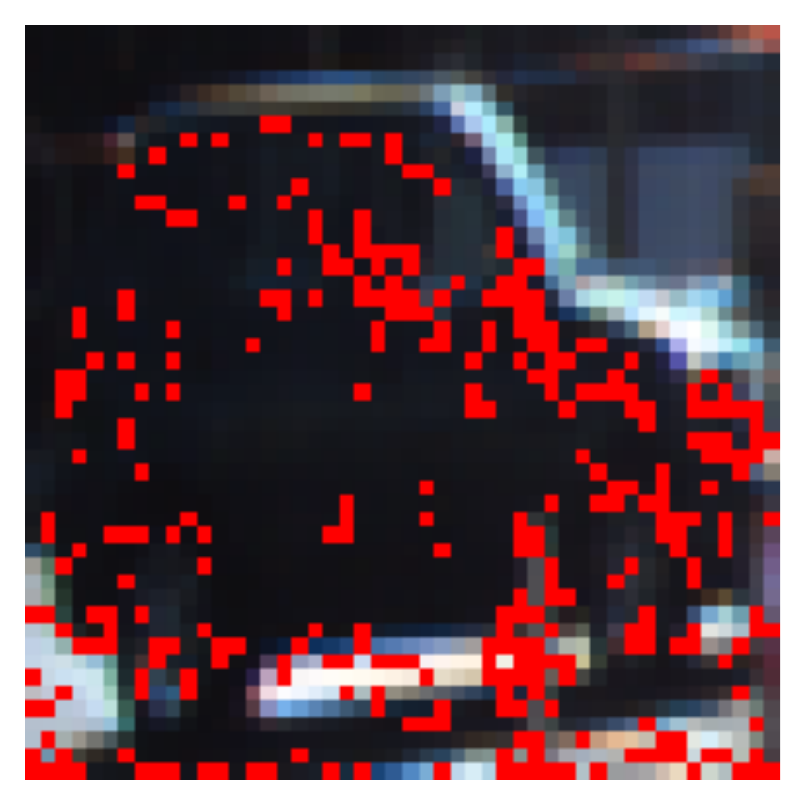}
\includegraphics[width=0.2\linewidth]{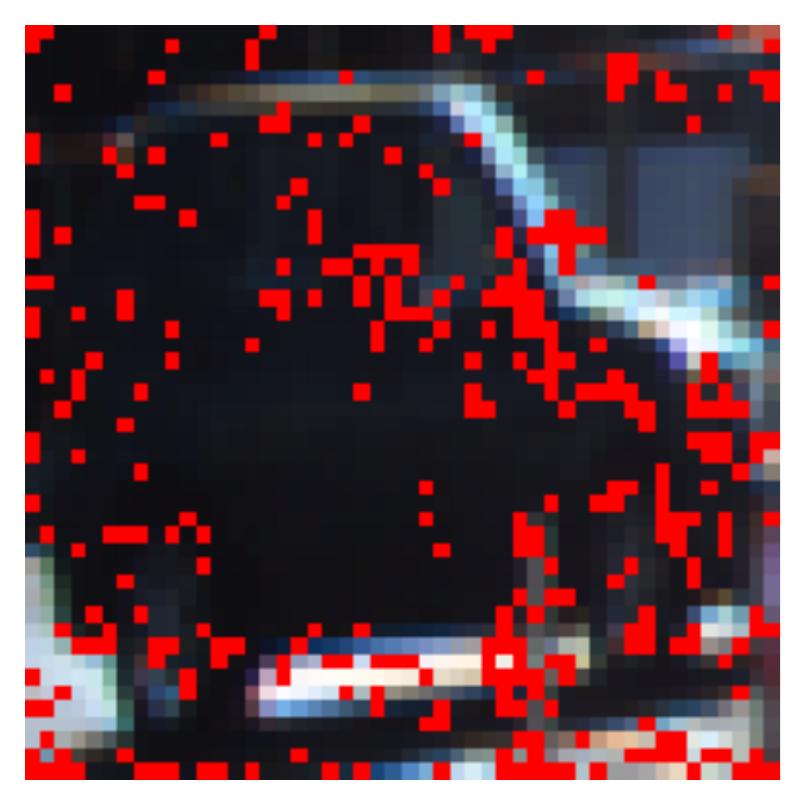}
\includegraphics[width=0.2\linewidth]{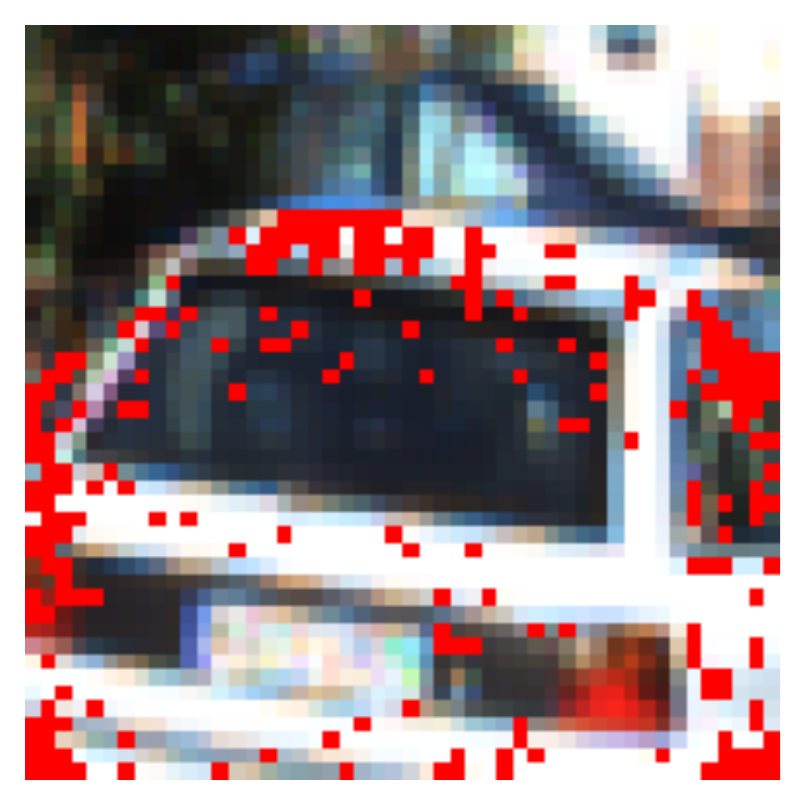}
\includegraphics[width=0.2\linewidth]{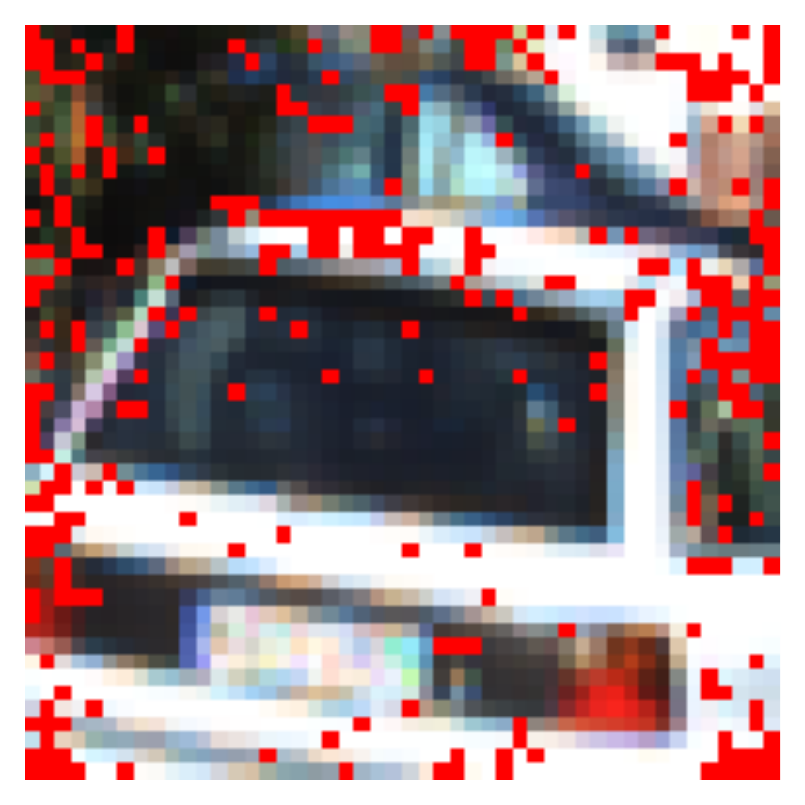}
\end{center}
   \caption{{\bf Qualitative comparison of max global pooling} on KITTI {\it validation} set. 
   The left/right image in each image pair marks the units activated by mask/standard global pooling.}
   \label{fig:maskpoolvis}
\end{figure}

\subsubsection{Instance assignment.}
We use a stand alone module to predict the `difficulty' of each instance, and assign it to its corresponding head network.
Tab.~\ref{table:assignment} shows the ablation study of this mechanism.
First, we can find that the accuracy of outputs increases with instance assignment.
Interestingly, considering that not all cases we can get the annotations of `difficulty', we use a simple alternative: using the distance from object to camera to represent the `difficulty' of objects (our default setting), and the threshold used in this experiment is (30, 50).
Experiment shows that this scheme get a similar performance as predicted difficulty levels.

\begin{table}[!ht]
\begin{center}
\caption{{\bf Ablation study of instance assignment} on KITTI {\it validation} set. Metrics are $AP_{3D}$ and $AP_{BEV}$ of the {\bf Car} category with 11 recall positions.}
\begin{tabular}{cccccccc}
\toprule
\multirow{2}{*}{assignment}& \multirow{2}{*}{switcher} & \multicolumn{3}{c}{$AP_{3D}$} & \multicolumn{3}{c}{$AP_{BEV}$} \\ 
\cmidrule{3-5} \cmidrule{6-8} 
~ & ~  & Easy & Moderate & Hard  & Easy & Moderate & Hard\\ 
\midrule
 - & -   & 33.7 & 21.5 & 19.2 & 42.5 & 28.2 & 23.5 \\
\checkmark &  difficulty & 34.7 & 22.1 & 19.5 & 44.1 & 29.0 & 24.2 \\
\checkmark &  distance & 35.1 & 22.0 & 19.6 & 44.4 & 29.1 & 24.1 \\
\bottomrule
\end{tabular}
\end{center}
\label{table:assignment}
\end{table}

\subsection{Comparing with state-of-the-art methods}

\begin{table}[!ht]
\caption{{\bf 3D detection performance} of the {\bf Car} category on KITTI dataset.
For {\it testing} set, only $AP|_{R_{40}}$ is provided by the official leaderboard.
For {\it validation} set, we report both $AP|_{R_{40}}$ and $AP|_{R_{11}}$ for better comparisons.
IoU threshold is set to 0.7.
$\ast$ indicates method is based on pseudo-LiDAR data.
Methods are ranked by {\it moderate} setting (same as KITTI leaderboard).
We highlight the best results in {\bf bold}.}
\begin{center}
\begin{tabular}{lccccccccc}
\toprule
\multirow{2}{*}{Method} & \multicolumn{3}{c}{testing ($AP|_{40}$)} & \multicolumn{3}{c}{validation$(AP|_{40})$} & \multicolumn{3}{c}{validation $(AP|_{11})$}\\ 
\cmidrule(r){2-4} \cmidrule(r){5-7} \cmidrule(r){8-10}
 ~ &  Easy & Mod. & Hard  & Easy & Mod. & Hard  & Easy & Mod. & Hard\\ 
\midrule
OFTNet~\cite{roddick2018orthographic} & 1.61 & 1.32 & 1.00 & - & - & -& 4.07 & 3.27 & 3.29 \\
FQNet~\cite{liu2019deep} & 2.77 & 1.51 & 1.01 & - & - & - & 5.98 & 5.50 & 4.75  \\
ROI-10D~\cite{Manhardt_2019_CVPR} & 4.32 & 2.02 & 1.46 & - & - & - & 10.25 & 6.39 & 6.18\\
GS3D~\cite{li2019gs3d}  & 4.47 & 2.90 & 2.47 & - & - & - & 13.46 & 10.97 & 10.38 \\
Shift R-CNN~\cite{naiden2019shift} & 6.88 & 3.87 & 2.83 & - & - & - & 13.84 & 11.29 & 11.08 \\
Multi-Fusion~\cite{Xu_2018_CVPR} & 7.08 & 5.18 & 4.68 & - & - & - & 22.03& 13.63& 11.60 \\ 
MonoGRNet~\cite{qin2019monogrnet} & 9.61 & 5.74 & 4.25 & - & - & - & 13.88 & 10.19 & 7.62 \\
Decoupled-3D* \cite{cai2020monocular}& 11.08 & 7.02 & 5.63 & - & - & - & 26.95 & 18.68 & 15.82 \\
MonoPSR~\cite{ku2018joint} & 10.76 & 7.25 & 5.85 & - & - & - & 12.75 & 11.48 & 8.59 \\
MonoPL* \cite{Weng_2019_ICCV_Workshops} & 10.76 & 7.50 & 6.10 & - & - & - & 31.5 & 21.00 & 17.50\\
SS3D~\cite{DBLP:journals/corr/abs-1906-08070} & 10.78 & 7.68 & 6.51 & - & - & - & 14.52 & 13.15 & 11.85 \\
MonoDIS~\cite{Simonelli_2019_ICCV} & 10.37 & 7.94 & 6.40 & 11.06 & 7.60 & 6.37 & 18.05 & 14.98 & 13.42\\
M3D-RPN~\cite{Brazil_2019_ICCV}  & 14.76 & 9.71 & 7.42 & - & - & -& 20.27 & 17.06 & 15.21 \\
PL-AVOD* \cite{Wang_2019_CVPR} & - & - & - & - & - & - & 19.5 & 17.2 & 16.2 \\
PL-FPointNet* \cite{Wang_2019_CVPR} & - & - & - & - & - & -& 28.2 & 18.5 & 16.4\\
AM3D* \cite{Ma_2019_ICCV}  & {\bf 16.50} & 10.74 & 9.52& 28.31 & 15.76 & 12.24 & 32.23 & 21.09 & 17.26  \\
PatchNet  & 15.68 & {\bf 11.12} & {\bf 10.17} & {\bf 31.6} & {\bf 16.8} & {\bf 13.8} & {\bf 35.1} & {\bf 22.0} & {\bf 19.6} \\
\bottomrule
\end{tabular}
\end{center}
\label{table:testset}
\end{table}

As shown in Tab.~\ref{table:testset}, we report our 3D detection results on the car category on KITTI dataset, where the proposed PatchNet ranks 1st among all published methods (ranked by {\it moderate} setting).
Overall, our method achieves superior result over other state-of-the-art methods across all settings except for {\it easy} level of {\it testing} set.
For instance, we outperform the current state-of-the-art AM3D~\cite{Ma_2019_ICCV} by {\bf 0.65/1.56/2.34} under $hard$ setting on the listed three metrics, which is the most challenging cases in the KITTI dataset.
Besides, the proposed method outperforms existing pseudo-LiDAR based approaches.
Note note we use the same depth estimator (DORN) as \cite{Ma_2019_ICCV,Wang_2019_CVPR,you2019pseudo,cai2020monocular} and the pipeline of proposed method is much simpler than pseudo-LiDAR based counterparts~\cite{cai2020monocular,you2019pseudo}.
This shows the effectiveness of our design.
We also observe that proposed model lags behind AM3D~\cite{Ma_2019_ICCV} under the {\it easy} setting on {\it testing} set.
This may be attributed to the differences of the 2D detectors.
We emphasize that {\it easy} split contains the least number of examples, so the performance of this setting is prone to fluctuations.
Also note that these three splits are containment relationships (e.g., {\it hard} split contains all instances belong to {\it easy} and {\it moderate} setting).

\subsection{Qualitative results}

\begin{figure}[t]
\begin{center}
\includegraphics[width=0.3\linewidth]{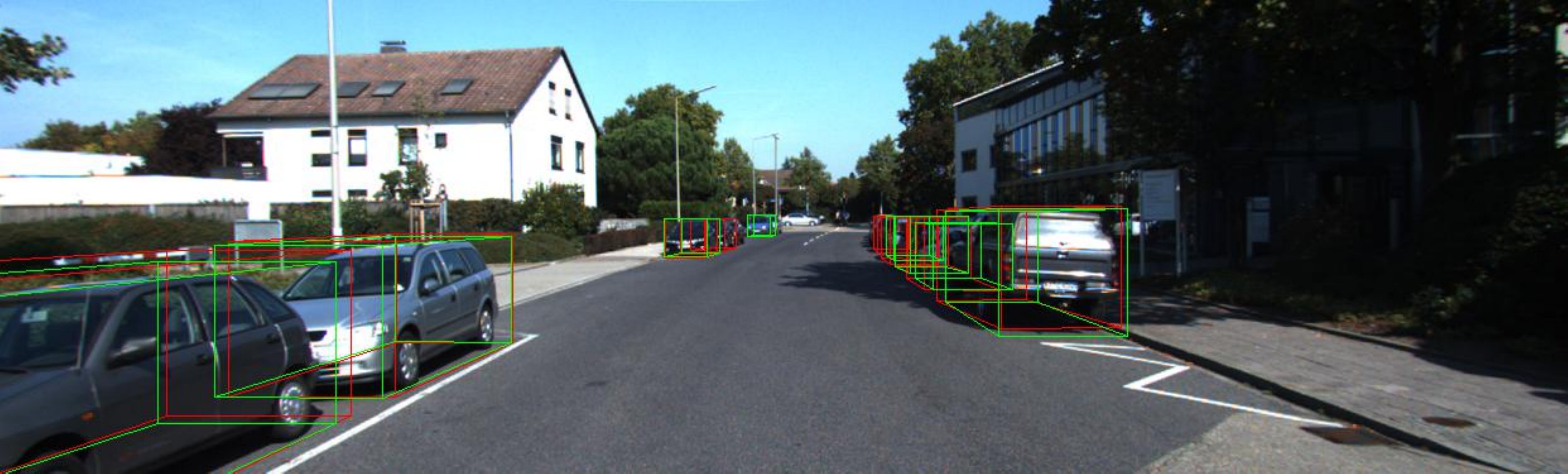}
\includegraphics[width=0.3\linewidth]{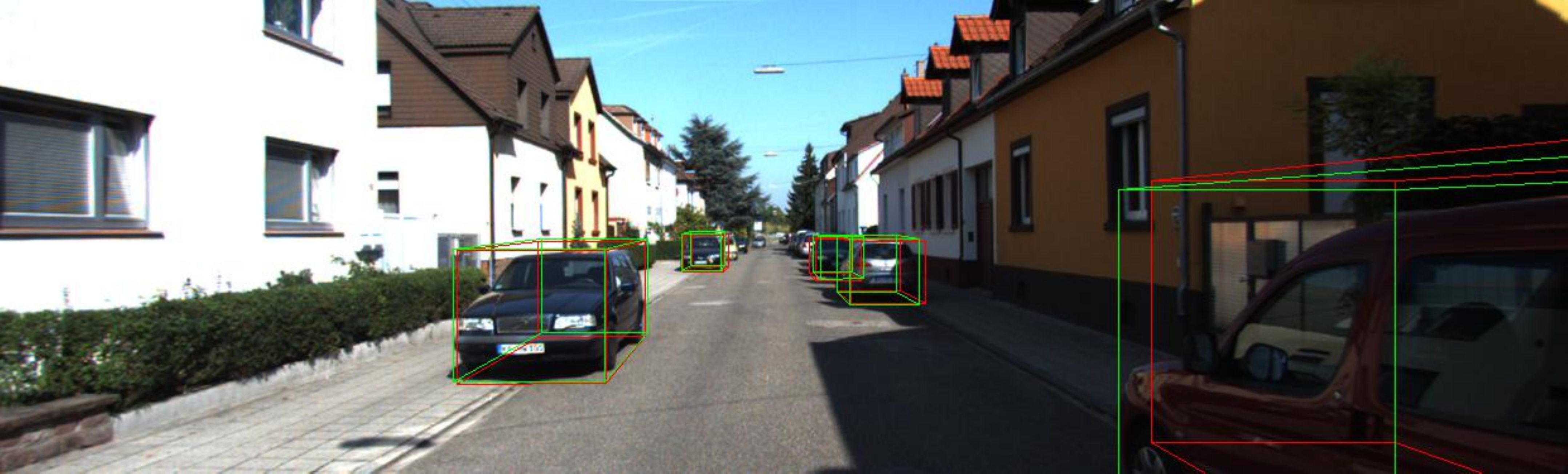}
\includegraphics[width=0.3\linewidth]{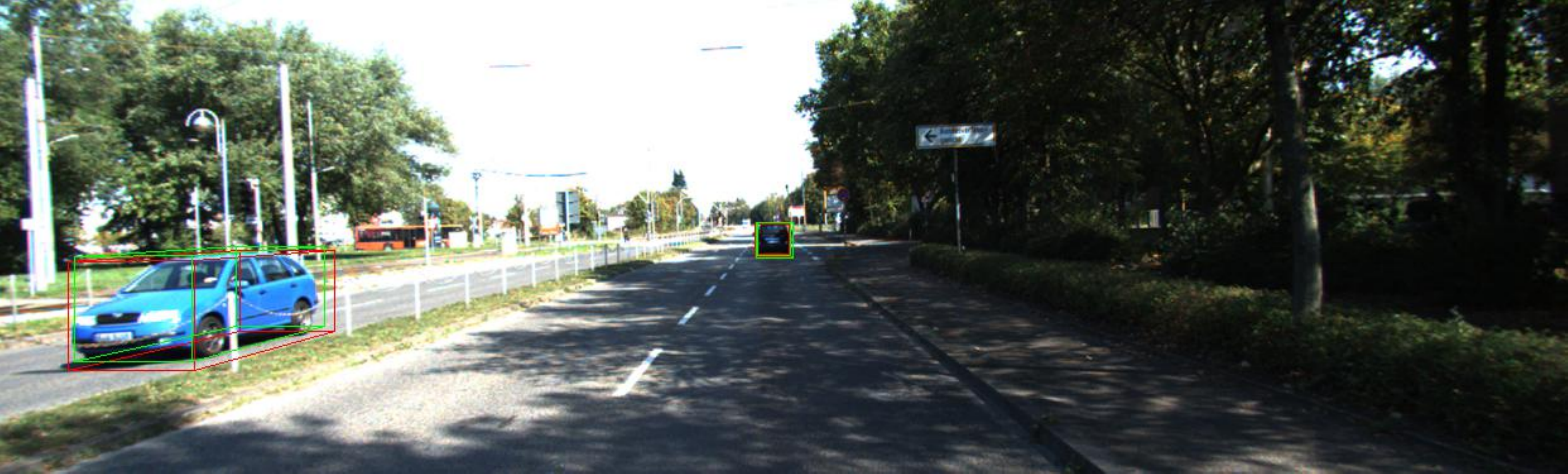}
\includegraphics[width=0.3\linewidth]{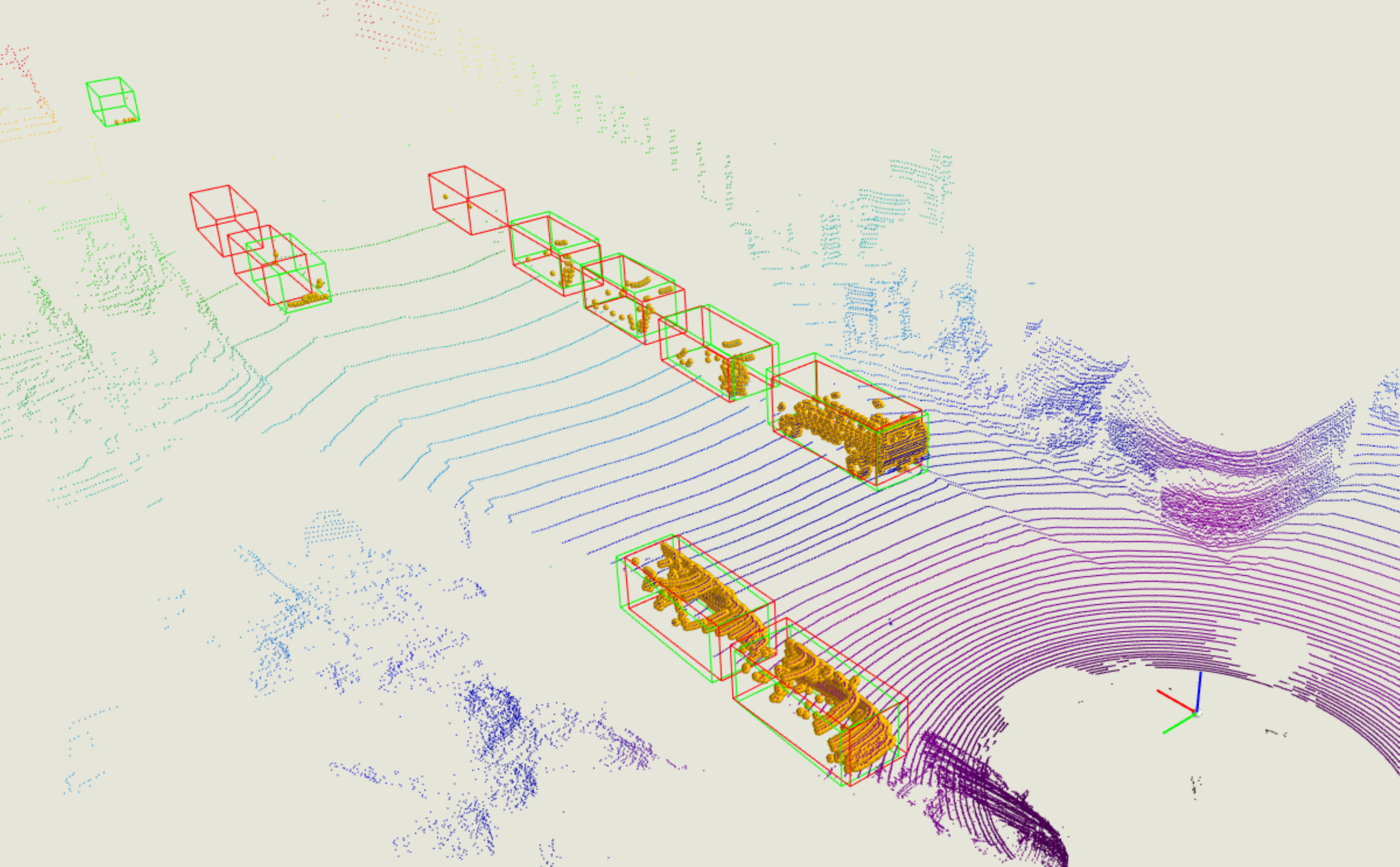}
\includegraphics[width=0.3\linewidth]{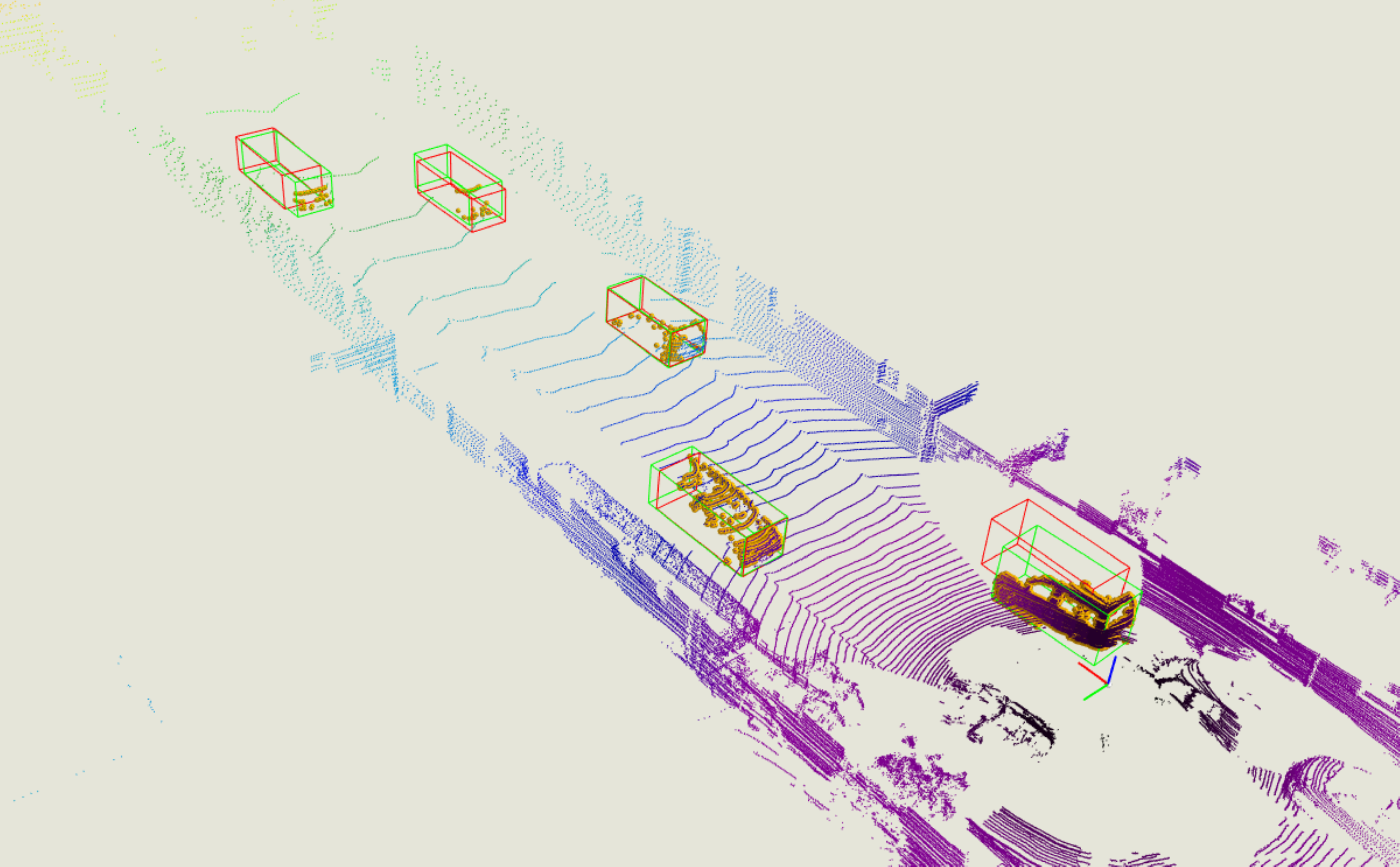}
\includegraphics[width=0.3\linewidth]{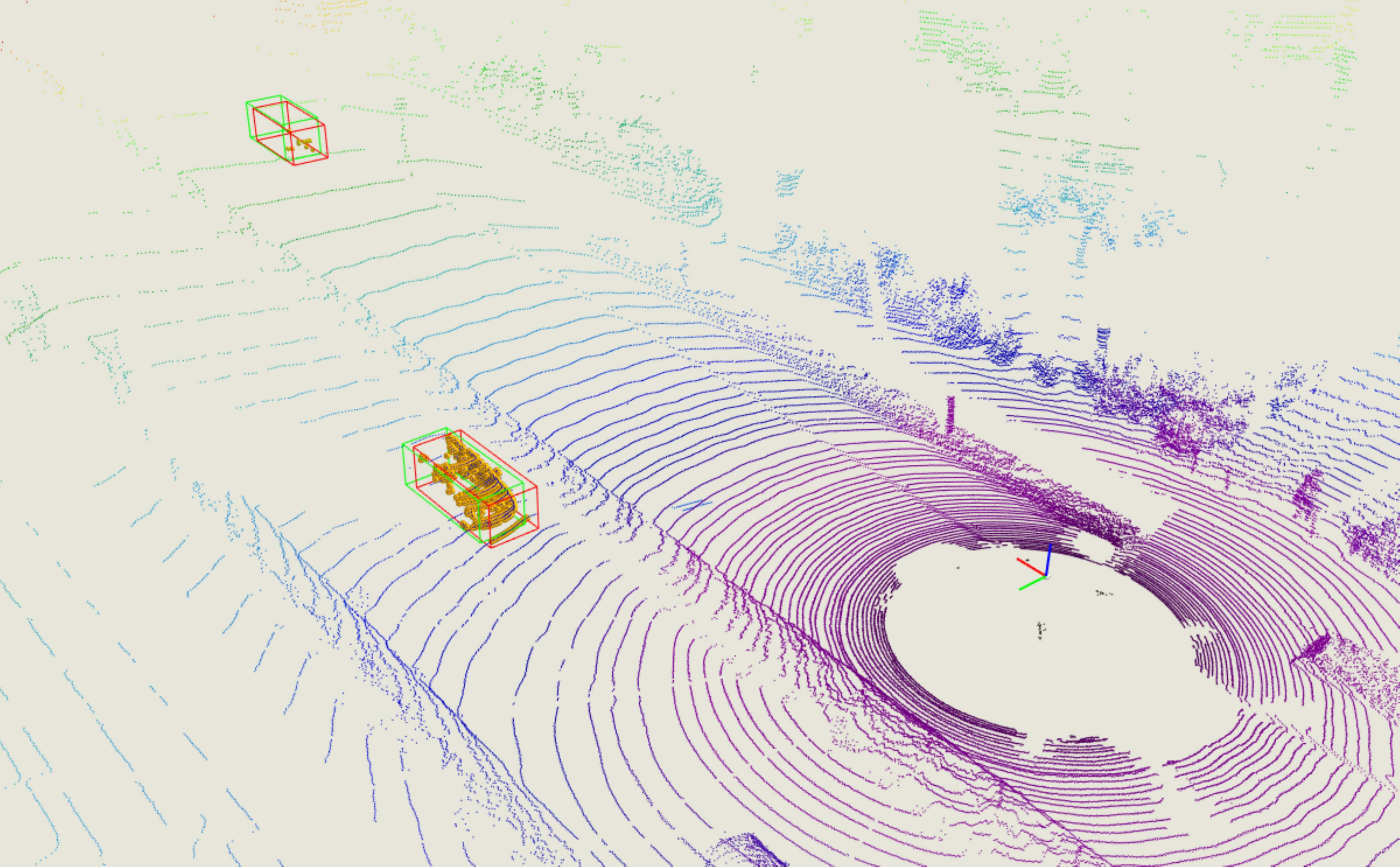}
\includegraphics[width=0.3\linewidth]{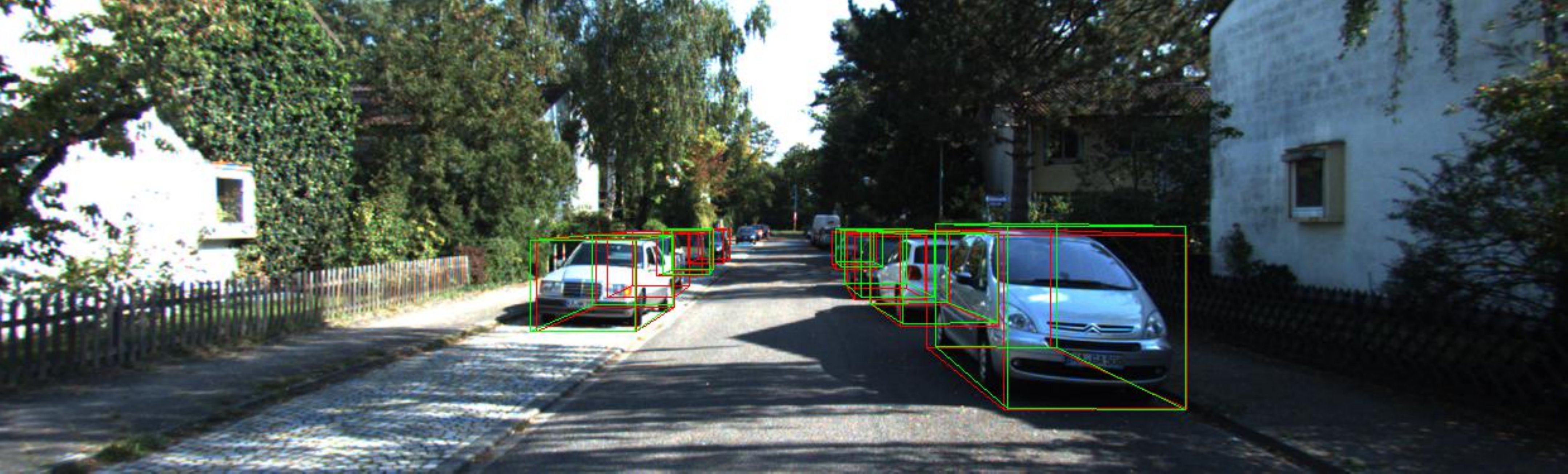}
\includegraphics[width=0.3\linewidth]{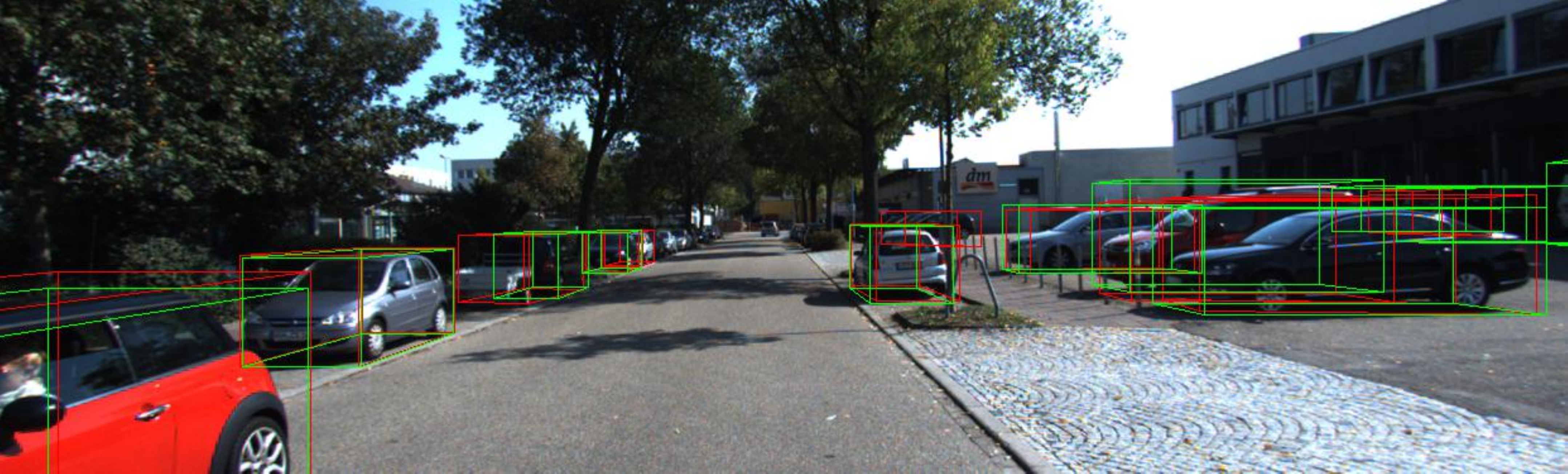}
\includegraphics[width=0.3\linewidth]{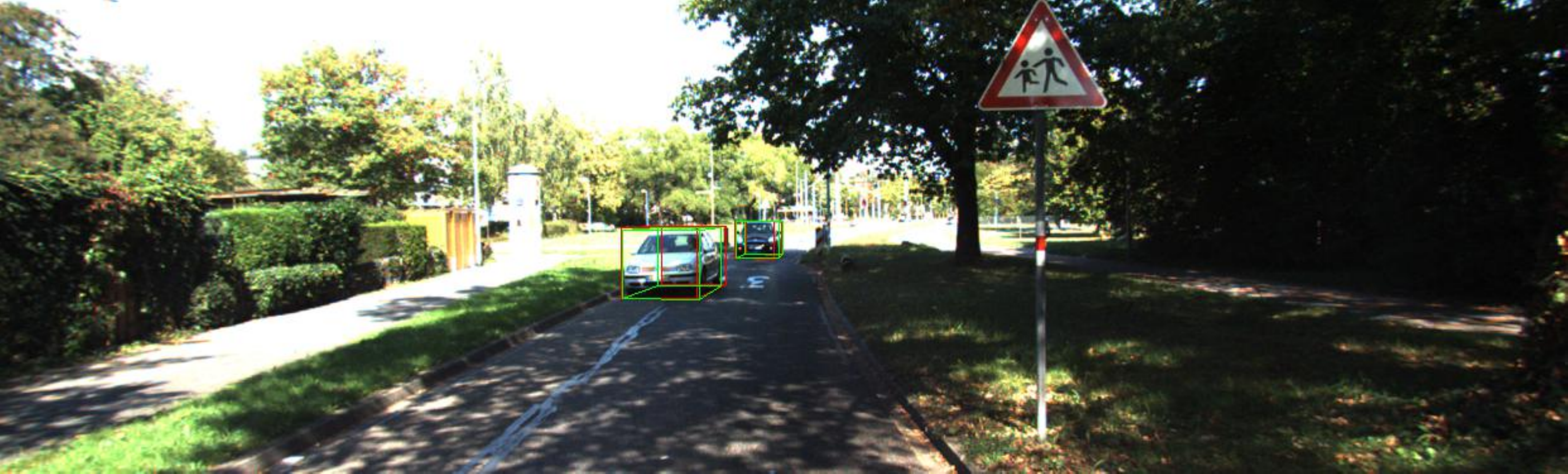}
\includegraphics[width=0.3\linewidth]{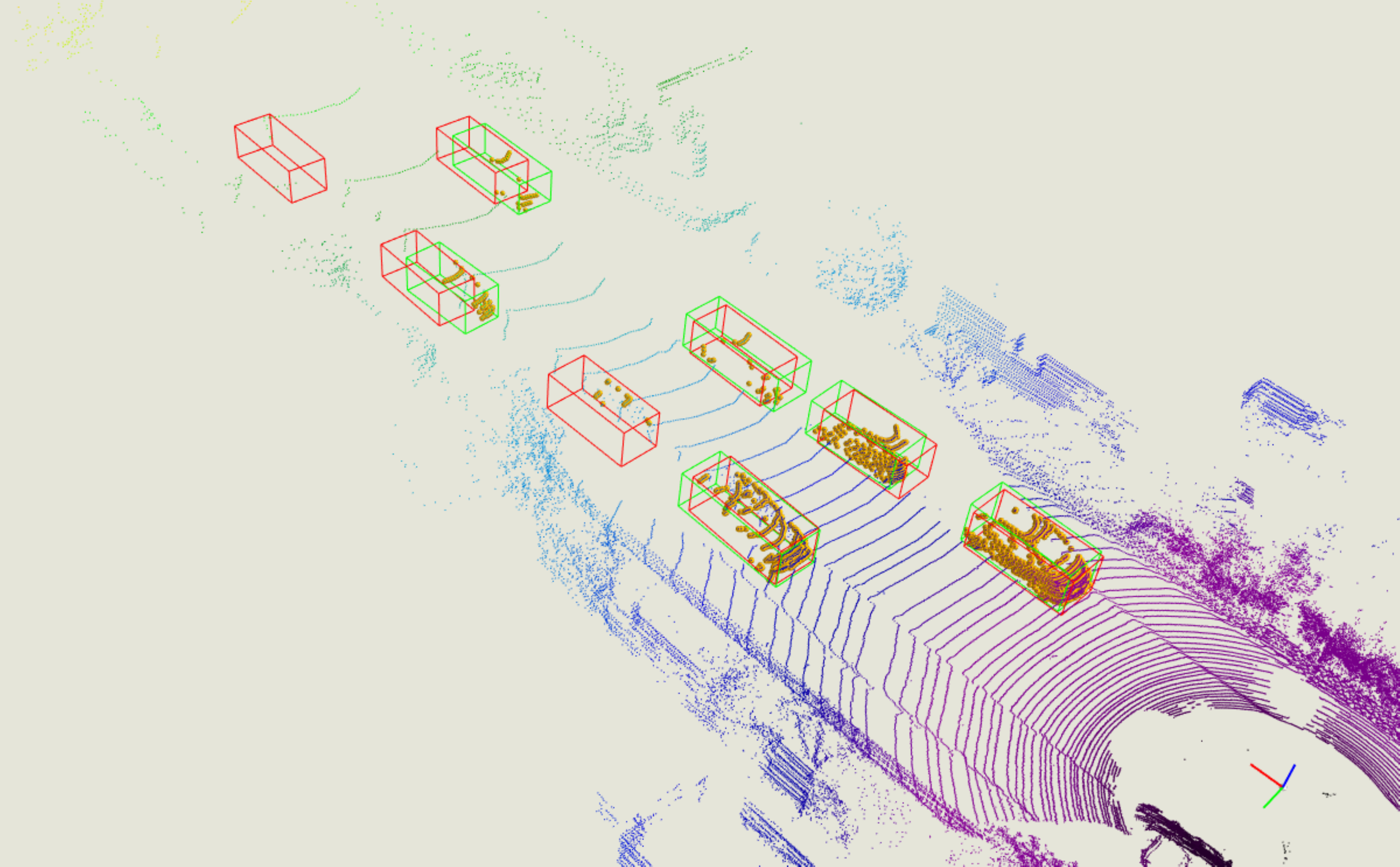}
\includegraphics[width=0.3\linewidth]{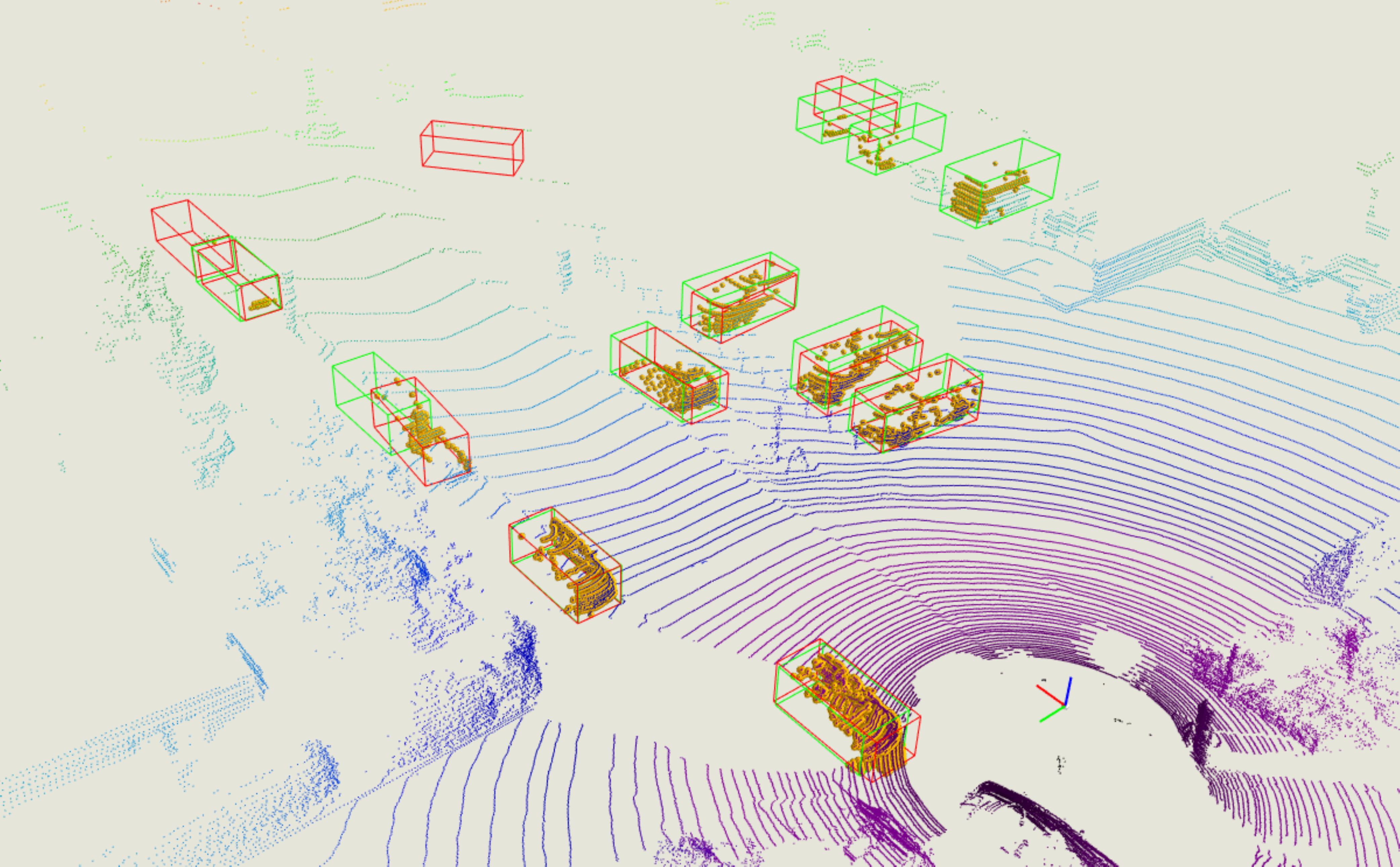}
\includegraphics[width=0.3\linewidth]{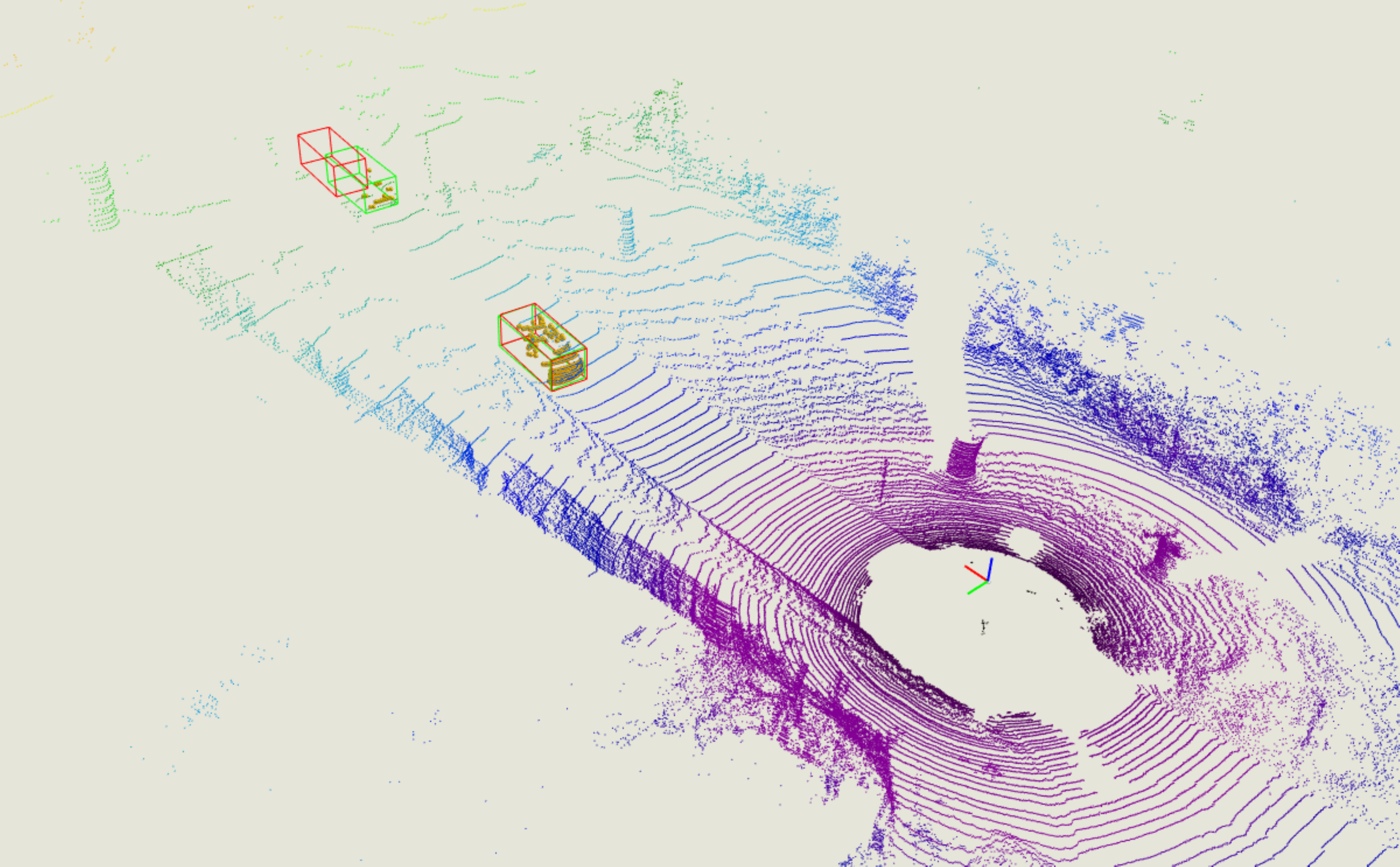}
\end{center}
   \caption{{\bf Qualitative results} on KITTI {\it validation} set. Red boxes represent our predictions, and green boxes come from ground truth. LiDAR signals are only used for visualization. Best viewed in color with zoom in.}
   \label{fig:vis}
\end{figure}

We visualize some representative outputs of our PatchNet model in Fig.~\ref{fig:vis}.
We can observe that for simple cases in reasonable distance, our model outputs remarkably accurate 3D bounding boxes.
Relatively, for distant objects, our estimates of their size and heading angle are still accurate, although it is difficult to determine its center.

On the other hand, we do observe several failure patterns, which indicate possible directions for future efforts.
First, our method often makes mistakes with truncated/occluded objects, and often manifests itself as inaccurate heading estimates.
Second, sometimes our 2D detector misses object due to strong occlusion, which will cause these samples to be ignored in subsequent steps.


\section{Conclusions}

In this paper, a novel network architecture, namely PatchNet, is proposed to explore the fundamental cause why pseudo-LiDAR representation based 3D detectors achieve promising performance.
Different from other works, we argue that the key factor is projecting the image coordinates to the world coordinates by the camera parameters,  rather than the point cloud representation itself.
More importantly, the world coordinate representation can be easily integrated into image representation, which means we can further boost the performance of 3D detector using more flexible and mature 2D CNN technologies.
Experimental results on KITTI dataset demonstrate our argument and show potential of image representation based 3D detector.
We hope these novel viewpoints provide insights to monocular/stereo 3D object detection community, and promote the development of new 2D CNN designs for image based 3D detection.


\section{Acknowledgement}
This work was supported by SenseTime, the Australian Research Council Grant  DP200103223, and Australian Medical Research Future Fund MRFAI000085.

\clearpage
\bibliographystyle{splncs04}
\bibliography{egbib}

\begin{thebibliography}{10}
\providecommand{\url}[1]{\texttt{#1}}
\providecommand{\urlprefix}{URL }
\providecommand{\doi}[1]{https://doi.org/#1}

\bibitem{Alhashim2018}
Alhashim, I., Wonka, P.: High quality monocular depth estimation via transfer
  learning. arXiv e-prints  \textbf{abs/1812.11941}, arXiv:1812.11941 (2018),
  \url{https://arxiv.org/abs/1812.11941}

\bibitem{Brazil_2019_ICCV}
Brazil, G., Liu, X.: M3d-rpn: Monocular 3d region proposal network for object
  detection. In: The IEEE International Conference on Computer Vision (ICCV)
  (October 2019)

\bibitem{cai2020monocular}
Cai, Y., Li, B., Jiao, Z., Li, H., Zeng, X., Wang, X.: Monocular 3d object
  detection with decoupled structured polygon estimation and height-guided
  depth estimation. arXiv preprint arXiv:2002.01619  (2020)

\bibitem{chabot2017deep}
Chabot, F., Chaouch, M., Rabarisoa, J., Teuliere, C., Chateau, T.: Deep manta:
  A coarse-to-fine many-task network for joint 2d and 3d vehicle analysis from
  monocular image. In: The IEEE Conference on Computer Vision and Pattern
  Recognition (CVPR). pp. 2040--2049 (2017)

\bibitem{chang2018pyramid}
Chang, J.R., Chen, Y.S.: Pyramid stereo matching network. In: Proceedings of
  the IEEE Conference on Computer Vision and Pattern Recognition. pp.
  5410--5418 (2018)

\bibitem{chen2016monocular}
Chen, X., Kundu, K., Zhang, Z., Ma, H., Fidler, S., Urtasun, R.: Monocular 3d
  object detection for autonomous driving. In: The IEEE Conference on Computer
  Vision and Pattern Recognition (CVPR). pp. 2147--2156 (2016)

\bibitem{chen20153d}
Chen, X., Kundu, K., Zhu, Y., Berneshawi, A.G., Ma, H., Fidler, S., Urtasun,
  R.: 3d object proposals for accurate object class detection. In: Advances in
  Neural Information Processing Systems. pp. 424--432 (2015)

\bibitem{Chen_2017_CVPR}
Chen, X., Ma, H., Wan, J., Li, B., Xia, T.: Multi-view 3d object detection
  network for autonomous driving. In: The IEEE Conference on Computer Vision
  and Pattern Recognition (CVPR) (July 2017)

\bibitem{dai2016r}
Dai, J., Li, Y., He, K., Sun, J.: R-fcn: Object detection via region-based
  fully convolutional networks. In: Advances in neural information processing
  systems. pp. 379--387 (2016)

\bibitem{fu2018deep}
Fu, H., Gong, M., Wang, C., Batmanghelich, K., Tao, D.: Deep ordinal regression
  network for monocular depth estimation. In: Proceedings of the IEEE
  Conference on Computer Vision and Pattern Recognition. pp. 2002--2011 (2018)

\bibitem{geiger2012we}
Geiger, A., Lenz, P., Urtasun, R.: Are we ready for autonomous driving? the
  kitti vision benchmark suite. In: 2012 IEEE Conference on Computer Vision and
  Pattern Recognition. pp. 3354--3361. IEEE (2012)

\bibitem{girshick2015fast}
Girshick, R.: Fast r-cnn. In: Proceedings of the IEEE international conference
  on computer vision. pp. 1440--1448 (2015)

\bibitem{godard2017unsupervised}
Godard, C., Mac~Aodha, O., Brostow, G.J.: Unsupervised monocular depth
  estimation with left-right consistency. In: Proceedings of the IEEE
  Conference on Computer Vision and Pattern Recognition. pp. 270--279 (2017)

\bibitem{he2017mask}
He, K., Gkioxari, G., Doll{\'a}r, P., Girshick, R.: Mask r-cnn. In: Proceedings
  of the IEEE international conference on computer vision. pp. 2961--2969
  (2017)

\bibitem{He_2016_CVPR}
He, K., Zhang, X., Ren, S., Sun, J.: Deep residual learning for image
  recognition. In: The IEEE Conference on Computer Vision and Pattern
  Recognition (CVPR) (June 2016)

\bibitem{Hu_2018_CVPR}
Hu, J., Shen, L., Sun, G.: Squeeze-and-excitation networks. In: The IEEE
  Conference on Computer Vision and Pattern Recognition (CVPR) (June 2018)

\bibitem{DBLP:journals/corr/abs-1906-08070}
Jörgensen, E., Zach, C., Kahl, F.: Monocular 3d object detection and box
  fitting trained end-to-end using intersection-over-union loss. CoRR
  \textbf{abs/1906.08070} (2019), \url{http://arxiv.org/abs/1906.08070}

\bibitem{ku2018joint}
Ku, J., Mozifian, M., Lee, J., Harakeh, A., Waslander, S.L.: Joint 3d proposal
  generation and object detection from view aggregation. In: 2018 IEEE/RSJ
  International Conference on Intelligent Robots and Systems (IROS). pp.~1--8.
  IEEE (2018)

\bibitem{li2019gs3d}
Li, B., Ouyang, W., Sheng, L., Zeng, X., Wang, X.: Gs3d: An efficient 3d object
  detection framework for autonomous driving. In: Proceedings of the IEEE
  Conference on Computer Vision and Pattern Recognition. pp. 1019--1028 (2019)

\bibitem{Li_2019_CVPR}
Li, P., Chen, X., Shen, S.: Stereo r-cnn based 3d object detection for
  autonomous driving. In: Proceedings of the IEEE/CVF Conference on Computer
  Vision and Pattern Recognition (CVPR) (June 2019)

\bibitem{Lin_2017_CVPR}
Lin, T.Y., Dollar, P., Girshick, R., He, K., Hariharan, B., Belongie, S.:
  Feature pyramid networks for object detection. In: Proceedings of the IEEE
  Conference on Computer Vision and Pattern Recognition (CVPR) (July 2017)

\bibitem{lin2017focal}
Lin, T.Y., Goyal, P., Girshick, R., He, K., Doll{\'a}r, P.: Focal loss for
  dense object detection. In: Proceedings of the IEEE international conference
  on computer vision. pp. 2980--2988 (2017)

\bibitem{liu2019deep}
Liu, L., Lu, J., Xu, C., Tian, Q., Zhou, J.: Deep fitting degree scoring
  network for monocular 3d object detection. In: Proceedings of the IEEE
  Conference on Computer Vision and Pattern Recognition. pp. 1057--1066 (2019)

\bibitem{Ma_2019_ICCV}
Ma, X., Wang, Z., Li, H., Zhang, P., Ouyang, W., Fan, X.: Accurate monocular 3d
  object detection via color-embedded 3d reconstruction for autonomous driving.
  In: The IEEE International Conference on Computer Vision (ICCV) (October
  2019)

\bibitem{Manhardt_2019_CVPR}
Manhardt, F., Kehl, W., Gaidon, A.: Roi-10d: Monocular lifting of 2d detection
  to 6d pose and metric shape. In: The IEEE Conference on Computer Vision and
  Pattern Recognition (CVPR) (June 2019)

\bibitem{mousavian20173d}
Mousavian, A., Anguelov, D., Flynn, J., Kosecka, J.: 3d bounding box estimation
  using deep learning and geometry. In: Proceedings of the IEEE Conference on
  Computer Vision and Pattern Recognition. pp. 7074--7082 (2017)

\bibitem{naiden2019shift}
Naiden, A., Paunescu, V., Kim, G., Jeon, B., Leordeanu, M.: Shift r-cnn: Deep
  monocular 3d object detection with closed-form geometric constraints. In:
  2019 IEEE International Conference on Image Processing (ICIP). pp. 61--65.
  IEEE (2019)

\bibitem{Qi_2018_CVPR}
Qi, C.R., Liu, W., Wu, C., Su, H., Guibas, L.J.: Frustum pointnets for 3d
  object detection from rgb-d data. In: The IEEE Conference on Computer Vision
  and Pattern Recognition (CVPR) (June 2018)

\bibitem{qi2017pointnet}
Qi, C.R., Su, H., Mo, K., Guibas, L.J.: Pointnet: Deep learning on point sets
  for 3d classification and segmentation. In: Proceedings of the IEEE
  conference on computer vision and pattern recognition. pp. 652--660 (2017)

\bibitem{qi2017pointnet++}
Qi, C.R., Yi, L., Su, H., Guibas, L.J.: Pointnet++: Deep hierarchical feature
  learning on point sets in a metric space. In: Advances in neural information
  processing systems. pp. 5099--5108 (2017)

\bibitem{qin2019monogrnet}
Qin, Z., Wang, J., Lu, Y.: Monogrnet: A geometric reasoning network for
  monocular 3d object localization. In: Proceedings of the AAAI Conference on
  Artificial Intelligence. vol.~33, pp. 8851--8858 (2019)

\bibitem{ren2015faster}
Ren, S., He, K., Girshick, R., Sun, J.: Faster r-cnn: Towards real-time object
  detection with region proposal networks. In: Advances in neural information
  processing systems. pp. 91--99 (2015)

\bibitem{roddick2018orthographic}
Roddick, T., Kendall, A., Cipolla, R.: Orthographic feature transform for
  monocular 3d object detection. arXiv preprint arXiv:1811.08188  (2018)

\bibitem{Simonelli_2019_ICCV}
Simonelli, A., Bulo, S.R., Porzi, L., Lopez-Antequera, M., Kontschieder, P.:
  Disentangling monocular 3d object detection. In: The IEEE International
  Conference on Computer Vision (ICCV) (October 2019)

\bibitem{Wang_2019_CVPR}
Wang, Y., Chao, W.L., Garg, D., Hariharan, B., Campbell, M., Weinberger, K.Q.:
  Pseudo-lidar from visual depth estimation: Bridging the gap in 3d object
  detection for autonomous driving. In: The IEEE Conference on Computer Vision
  and Pattern Recognition (CVPR) (June 2019)

\bibitem{Weng_2019_ICCV_Workshops}
Weng, X., Kitani, K.: Monocular 3d object detection with pseudo-lidar point
  cloud. In: IEEE International Conference on Computer Vision (ICCV) Workshops
  (Oct 2019)

\bibitem{xie2017aggregated}
Xie, S., Girshick, R., Doll{\'a}r, P., Tu, Z., He, K.: Aggregated residual
  transformations for deep neural networks. In: Proceedings of the IEEE
  conference on computer vision and pattern recognition. pp. 1492--1500 (2017)

\bibitem{Xu_2018_CVPR}
Xu, B., Chen, Z.: Multi-level fusion based 3d object detection from monocular
  images. In: The IEEE Conference on Computer Vision and Pattern Recognition
  (CVPR) (June 2018)

\bibitem{you2019pseudo}
You, Y., Wang, Y., Chao, W.L., Garg, D., Pleiss, G., Hariharan, B., Campbell,
  M., Weinberger, K.Q.: Pseudo-lidar++: Accurate depth for 3d object detection
  in autonomous driving. arXiv preprint arXiv:1906.06310  (2019)

\bibitem{zhou2020cheaper}
Zhou, D., Zhou, X., Zhang, H., Yi, S., Ouyang, W.: Cheaper pre-training lunch:
  An efficient paradigm for object detection. arXiv preprint arXiv:2004.12178
  (2020)

\end{thebibliography}


\clearpage
\setcounter{section}{0}
\renewcommand\thesection{\Alph{section}} 
\section{Overview}
This document provides additional analysis and extra experiments to the main paper.
Specifically, in Sec.~\ref{sec:runtime}, we analyse the latency of proposed method and compare it with some pseudo-LiDAR based methods. 
Sec.~\ref{sec:stereo} shows the performance of stereo images while Sec. \ref{sec:pedcyc} gives the results of Pedestrian and Cyclist detection. 
Finally, Sec.~\ref{sec:vis} presents more visualization examples.

\section{Runtime Analysis}
\label{sec:runtime}

In this section, we will analyze the latency of our PatchNet and compare it with some existing methods~\cite{Ma_2019_ICCV,Wang_2019_CVPR,Weng_2019_ICCV_Workshops} based on pseudo-LiDAR representation. 
In general, all the four methods can be divided into three main stages. 
In ours designs, the processing flows of PatchNet-vanilla and pseudo-LiDAR are the same, but the representations of inputs are different. 
So the runtime of these two methods are almost the same, which is shown as follow (tested on a single 1080 GPU):

\begin{table}
\caption{Runtime of PachNet-vanilla and pseudo-LiDAR.}
\begin{center}
\begin{tabular}{ccc}  
\toprule   
2D detection & Depth estimation & 3D detection \\ 
\midrule   
60ms &  400ms & 28ms\\
\bottomrule  
\end{tabular}
\end{center}
\label{table:runtim1}
\end{table}

PatchNet shares the same 2D detector and depth estimator (note the runtime of different depth estimators varies greatly, see
\href{http://www.cvlibs.net/datasets/kitti/eval_depth.php?benchmark=depth_prediction}{{\bf \underline{KITTI Benchmark}}} for details), and we show its runtime of 3D detection stage for different backbone models as follows:

\begin{table}
\caption{Runtime of PatchNet in 3D detection stage.}

\begin{center}
\begin{tabular}{ccccc}  
\toprule   
Backbone & PointNet-18 & ResNet-18 & ResNeXt-18 & SE-ResNet-18 \\ 
\midrule   
runtime & 12ms & 23ms & 18ms & 26ms\\
\bottomrule  
\end{tabular}
\end{center}
\label{table:runtim2}
\end{table}

Although we add some extra operations in PatchNet, the runtime of the baseline model (PointNet-18) is 12ms while the runtime of PatchNet-vanilla is 28ms. 
This is mainly because we remove the foreground segmentation net and use a dynamic threshold to segment the foreground, which can save about 18ms. 
For the best backbone, the runtime is only 26ms, which has similar runtime of pseudo-LiDAR for 3d detection.

Besides, although PatchNet and \cite{Ma_2019_ICCV} use the same segmentation method, \cite{Ma_2019_ICCV} add another ResNet-34 to extract image features. For \cite{Weng_2019_ICCV_Workshops}, it adds a 2D instance segmentation net, which will bring lots of computing overhead (e.g., about 200ms for Mask RCNN~\cite{he2017mask}).

In summary, PatchNet is more efficient than \cite{Ma_2019_ICCV,Weng_2019_ICCV_Workshops} and has the similar run time as \cite{Wang_2019_CVPR}.

\section{Stereo Images}
\label{sec:stereo}
Pseudo-LiDAR representation is also widely used in the field of stereo 3D detection task. 
In order to verify that the proposed method is still work with binocular images, we replace the monocular depth maps with the stereo ones (we use PSMNet~\cite{chang2018pyramid} as our stereo depth estimator and get the pre-trained model from \cite{Wang_2019_CVPR}) and test the performance on KITTI {\it validation} set using $AP|_{R_{11}}$ for better comparison with previous works. 
As shown in the Tab.~\ref{table:stereo} , PatchNet-vanilla has almost the same accuracy as pseudo-LiDAR, while PatchNet achieves better performances.
We also report the $AP|_{R_{40}}$ for reference.

\begin{table}
\caption{{\bf Stereo 3D detection performance} of the {\bf Car} category on KITTI {\it validation} dataset. IoU threshold is set to 0.7. We highlight the best results in {\bf bold}.}
\begin{center}
\begin{tabular}{lcccccc}
\toprule
\multirow{2}{*}{Method} & \multicolumn{3}{c}{3D Detection} & \multicolumn{3}{c}{BEV Detection} \\ 
\cmidrule(r){2-4}  \cmidrule(r){5-7}
 ~  & Easy & Moderate & Hard  & Easy & Moderate & Hard\\ 
\midrule
3DOP~\cite{chen20153d}   & 6.55 & 5.07 & 4.10 & 12.63 & 9.49 & 7.59 \\
Multi-Fusion~\cite{Xu_2018_CVPR}   & - & 9.80 & - & - & 19.54 & - \\
Stereo-RCNN~\cite{Li_2019_CVPR}   & 54.1 & 36.7 & 31.1 & 68.5 & 48.3 & 41.5 \\
Pseudo-LiDAR~\cite{Wang_2019_CVPR}  & 59.4 & 39.8 & 33.5 & 72.8 & 51.8 &  44.0 \\
PatchNet-vanilla & 60.8 & 40.1 & 33.6 & 72.7 & 51.2 & 43.8 \\
PatchNet   & {\bf 65.9} & {\bf 42.5} & {\bf 38.5} & {\bf 74.5} & {\bf 52.9} & {\bf 44.8} \\
\midrule
PatchNet-vanilla$@AP|_{R_{40}}$ & 61.4 & 37.6 & 31.6 & 73.5 & 49.8 & 41.7 \\
PatchNet$@AP|_{R_{40}}$   & {\bf 66.0} & {\bf 41.1} & {\bf 34.6} & {\bf 76.8} & {\bf 52.8} & {\bf 44.3} \\

\bottomrule
\end{tabular}
\end{center}
\label{table:stereo}
\end{table}

\section{Pedestrian and Cyclist}
\label{sec:pedcyc}

For better comparison, we also report {\bf Pedestrian/Cyclist} detection performance for 3D detection task on KITTI {\it validation} set in this part. 
Specifically, we conduct these experiments using both monocular and stereo images with $AP|_{R_{11}}$ as metric.
It can be seen from Tab.~\ref{table:pedcyc} that the proposed model also get better performance than \cite{Wang_2019_CVPR} with each setting. 
Note that results of pseudo-LiDAR are evaluated by ourselves using its official code, since pseudo-LiDAR did not provide Pedestrian/Cyclist detection results for monocular images. 

Besides, the accuracy of {\bf Pedestrian/Cyclist} detection fluctuate greatly compared with {\bf Car} detection.
This fluctuation of performance is mainly caused by insufficient training samples (there are only 2,207/734 training samples for {\bf Pedestrian/Cyclist} in KITTI {\it training} set, while it provides 14,357 {\bf Car} instances).
This problem can be reduced by introducing more training data or more effective data augmentation strategies.

\begin{table}
\caption{{\bf 3D detection performance} of the {\bf Pedestrian/Cyclist} category on KITTI {\it validation} dataset. Metric is $AP|_{R_{11}}$ and IoU threshold is set to 0.5. We highlight the best results in {\bf bold}.}
\begin{center}
\begin{tabular}{lccccccc}
\toprule
\multirow{2}{*}{Method} & \multirow{2}{*}{Category} & \multicolumn{3}{c}{Monocular} & \multicolumn{3}{c}{Stereo} \\ 
\cmidrule(r){3-5}  \cmidrule(r){6-8}
 ~ & ~ & Easy & Moderate & Hard  & Easy & Moderate & Hard\\ 
\midrule
Pseudo-LiDAR~\cite{Wang_2019_CVPR} & Pedestrian  & 7.32 & 6.19 & 5.64 & 33.8 & 27.4 & 24.0 \\
PatchNet & Pedestrian &  {\bf 9.82} &  {\bf 7.86} &  {\bf 6.84} & {\bf 38.8} &  {\bf 30.1} &  {\bf 26.5}  \\

Pseudo-LiDAR~\cite{Wang_2019_CVPR} & Cyclist & 5.49 & 3.85 & 3.82 & 41.3 & 25.2 & 24.9 \\
PatchNet & Cyclist & {\bf 8.14} & {\bf 4.84} & {\bf 4.62} & {\bf 46.8} & {\bf 29.0} & {\bf 26.8} \\

\bottomrule
\end{tabular}
\end{center}
\label{table:pedcyc}
\end{table}

\section{More Qualitative Examples}
\label{sec:vis}

\begin{figure}[h]
\begin{center}
\includegraphics[width=0.47\linewidth]{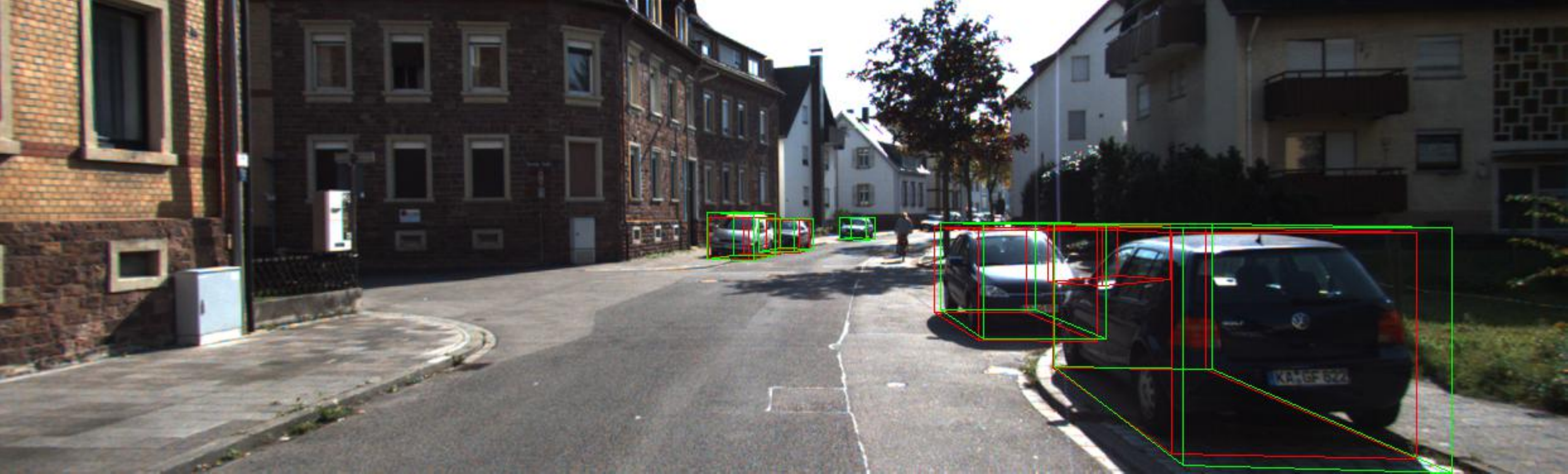}
\includegraphics[width=0.47\linewidth]{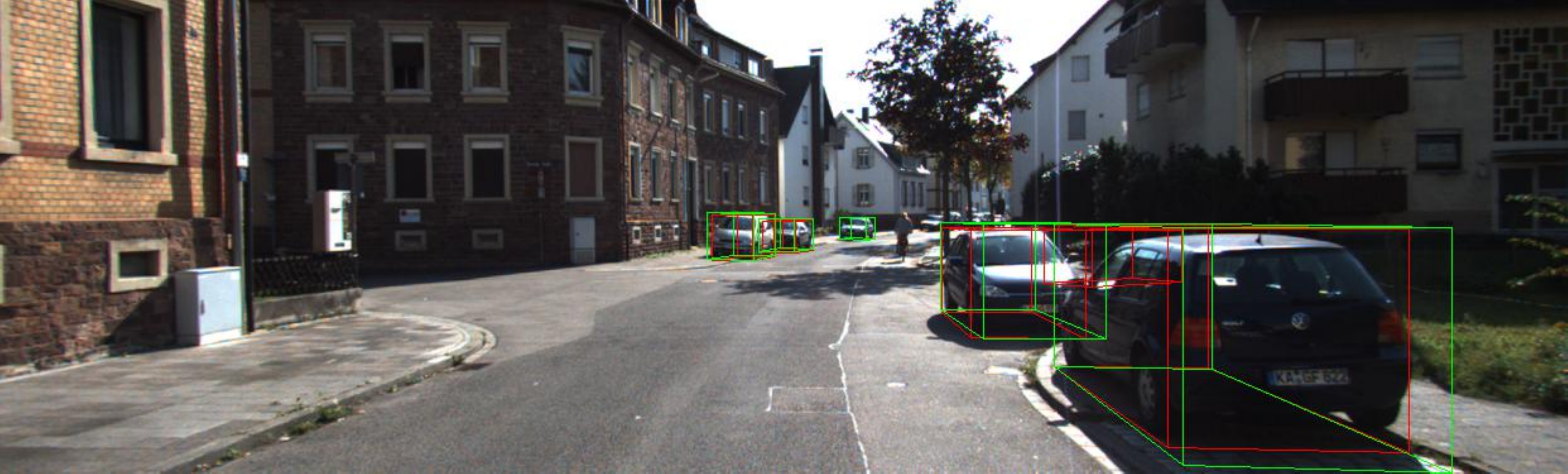}
\includegraphics[width=0.47\linewidth]{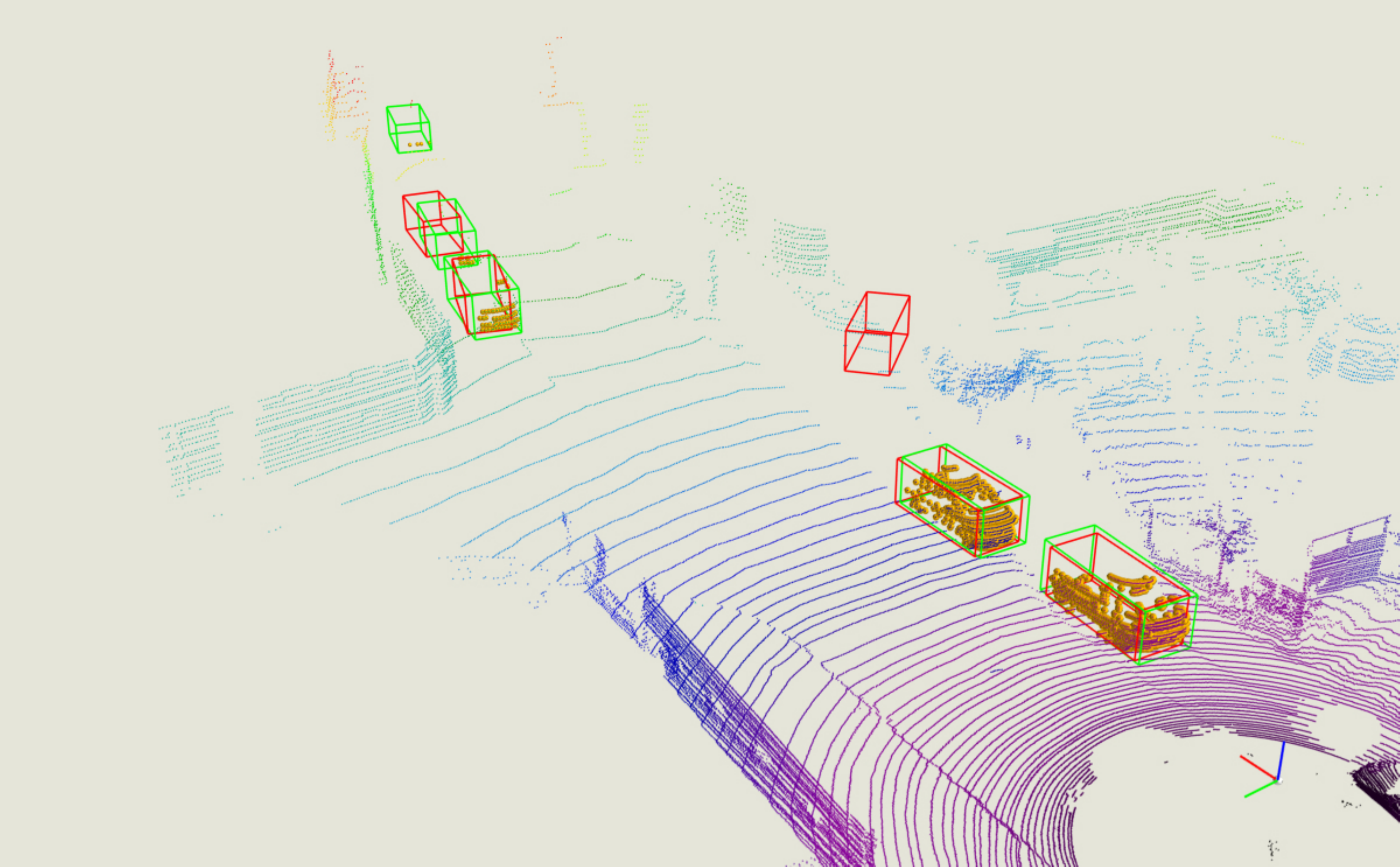}
\includegraphics[width=0.47\linewidth]{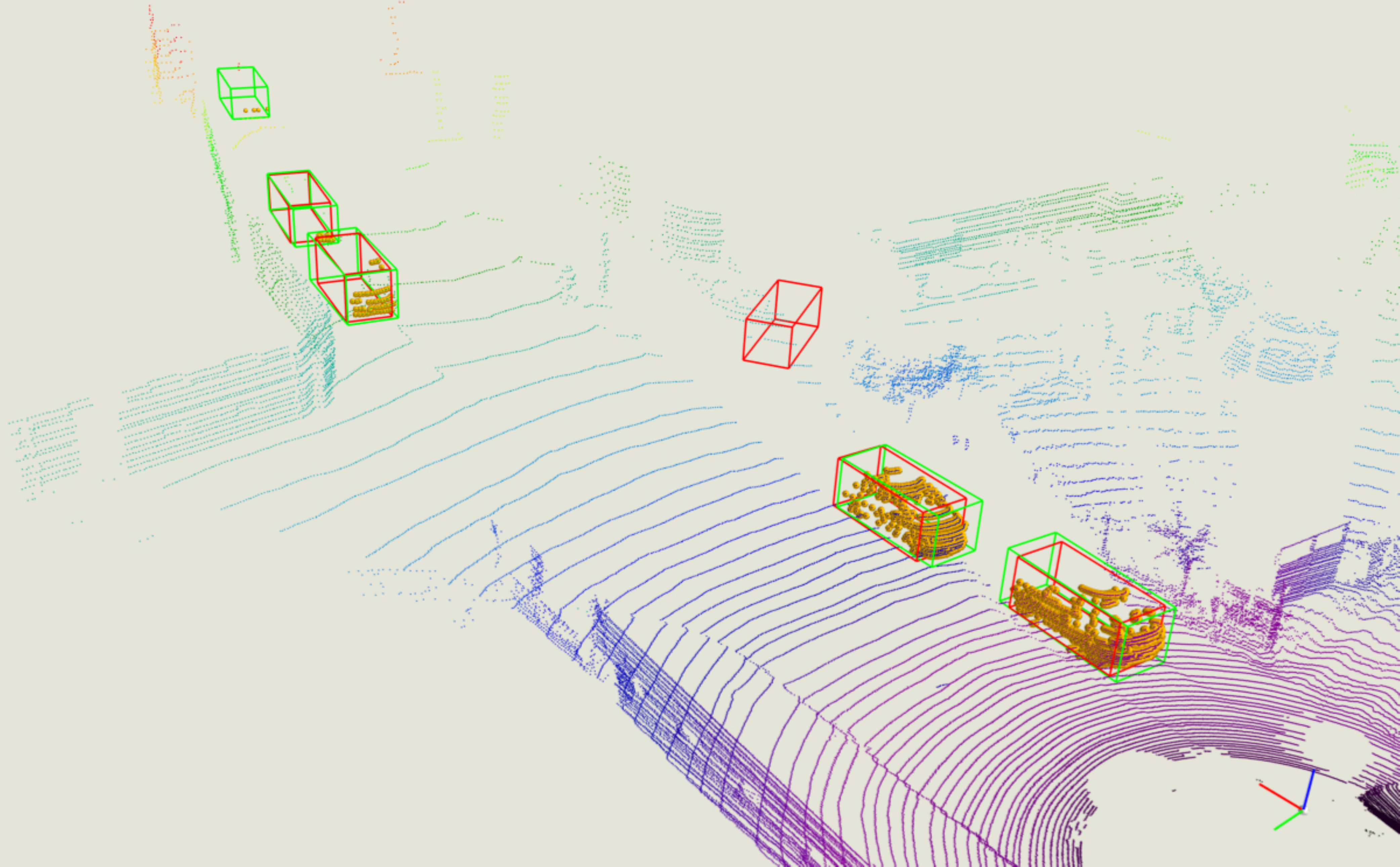}
\includegraphics[width=0.47\linewidth]{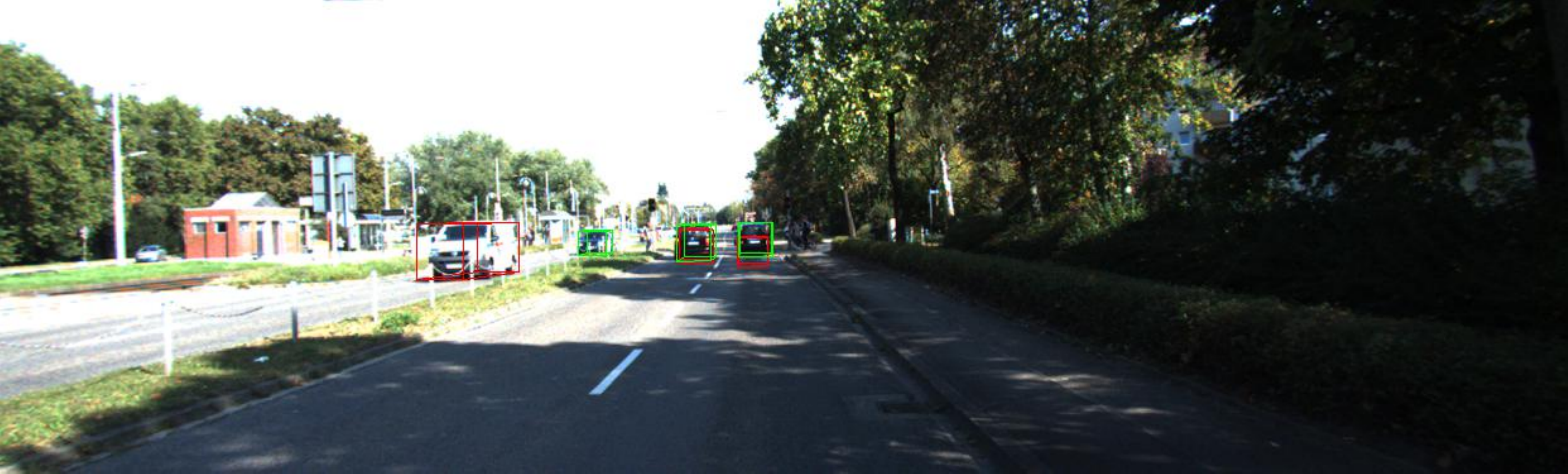}
\includegraphics[width=0.47\linewidth]{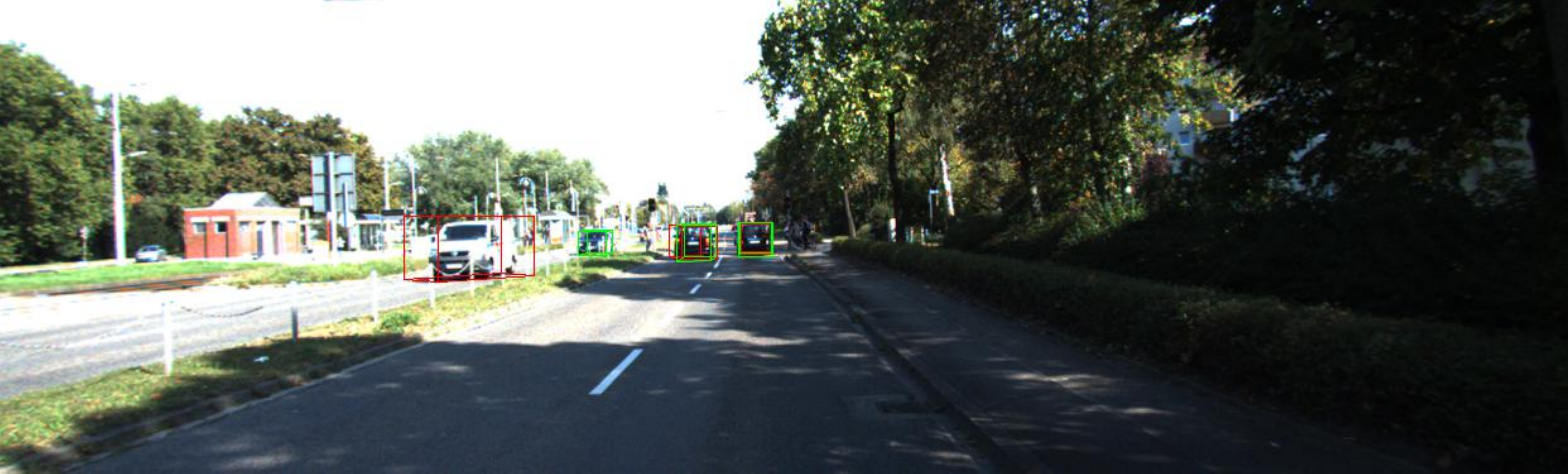}
\includegraphics[width=0.47\linewidth]{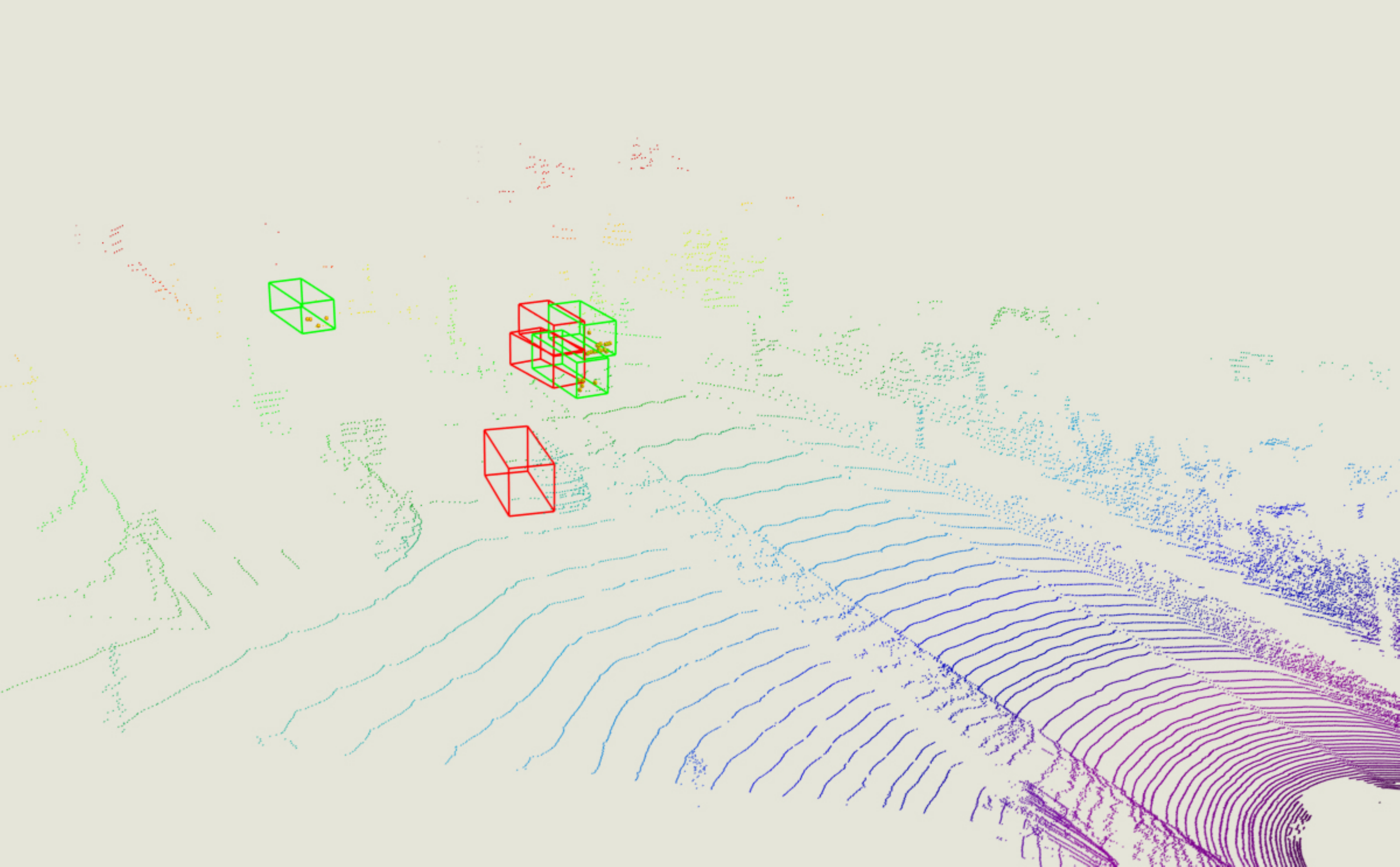}
\includegraphics[width=0.47\linewidth]{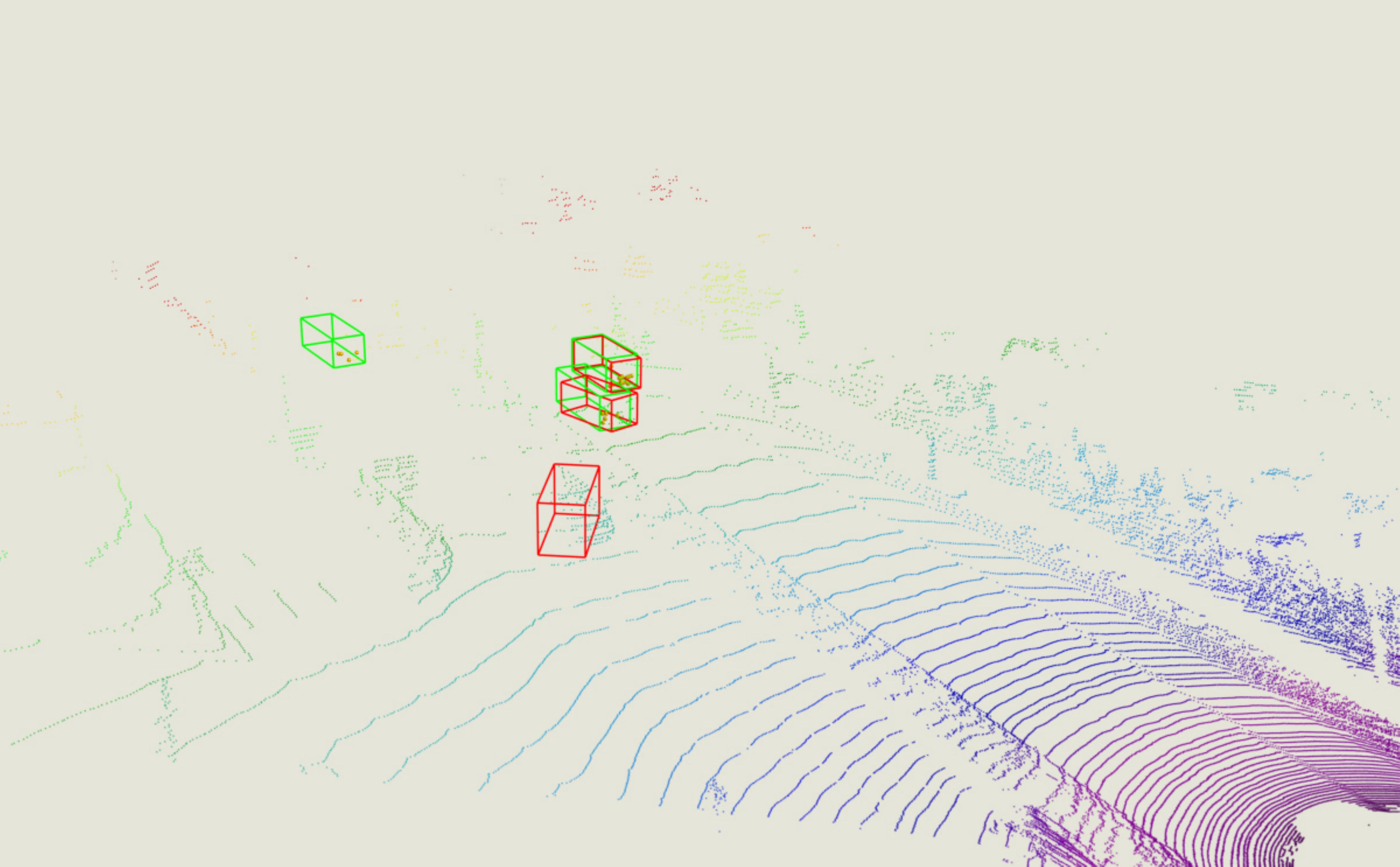}
\end{center}
   \caption{{\bf Qualitative results} on KITTI {\it validation} set.{\it Left}: monocular detection results. {\it Right}: stereo detection results.Red boxes represent our predictions, and green boxes come from ground truth. LiDAR signals are only used for visualization. Best viewed in color with zoom in.}
   \label{fig:vis2}
\end{figure}

In this part, we compare the monocular images and stereo pairs by some representitive qualitative results in Fig.~\ref{fig:vis2}.
First, we can find that stereo images can detect objects more accurately, which is generally reflected to the better depth estimation, instead of size or heading estimation.
Then, for most of close range objects, in terms of visual experience, monocular images are not inferior to stereo images (although there are still some failure case among those instances).


\end{document}